\documentclass[10pt,journal,compsoc]{IEEEtran}
\ifCLASSOPTIONcompsoc
  \usepackage[nocompress]{cite}
\else
  \usepackage{cite}
\fi
\ifCLASSINFOpdf
\else
\fi
\usepackage[margin=1in]{geometry} 
\usepackage{booktabs}
\usepackage{multirow}
\usepackage{caption,fixltx2e}
\usepackage[flushleft]{threeparttable}
\usepackage{subcaption}
\usepackage[flushleft]{threeparttable}
\usepackage{amsmath}
\usepackage{amsthm}
\usepackage{xspace}
\usepackage{amssymb}
\usepackage{graphicx}
\usepackage{lipsum}
\usepackage{float}
\usepackage{makecell}
\usepackage{lipsum,graphicx,multicol}
\usepackage{hyperref}
\hypersetup{
     colorlinks=true,
     linkcolor=blue,
     filecolor=blue,
     citecolor = black,      
     urlcolor=cyan,
     }
\usepackage{color,soul}
\usepackage{longtable}
\usepackage{mathrsfs}

\newcommand{\TopEvent}{\ensuremath{TE}\xspace}
\newcommand{\FTSize}{\phi_s \xspace}
\newcommand{\ErrorData}{\phi_d \xspace}
\newcommand{\ErrorMCSs}{\phi_c \xspace}

\begin{document}
%
\title{Automatic inference of fault tree models via multi-objective evolutionary algorithms}
%
%
%
%

\author{Lisandro A.~Jimenez-Roa,
        Tom~Heskes,
        Tiedo~Tinga,
        and~Marielle~Stoelinga
\IEEEcompsocitemizethanks{\IEEEcompsocthanksitem Lisandro A. Jimenez-Roa is with the Faculty of Electrical Engineering, Mathematics and Computer Science (EEMCS), University of Twente, Drienerlolaan 5, 7522 NB Enschede, The Netherlands.
\protect\\
E-mail: l.jimenezroa@utwente.nl
\IEEEcompsocthanksitem Tiedo Tinga is with the Faculty of Engineering Technology (ET), University of Twente, Drienerlolaan 5, 7522 NB Enschede, The Netherlands\protect\\
E-mail: t.tinga@utwente.nl
\IEEEcompsocthanksitem Tom Heskes is with the Institute for Computing and Information Sciences (iCIS), Radboud University Nijmegen, 6525 EC Nijmegen, The Netherlands.\protect\\
E-mail: Tom.Heskes@ru.nl
\IEEEcompsocthanksitem Marielle Stoelinga is with the Faculty of Electrical Engineering, Mathematics and Computer Science (EEMCS), University of Twente, and the Department of Software Science, Radboud University Nijmegen, The Netherlands.\protect\\
E-mail: m.i.a.stoelinga@utwente.nl}
\thanks{Manuscript received September 24, 2021; revised -, 2021.}}

%
%

\markboth{IEEE transactions on dependable and secure computing\, ~Vol.~XX, No.~X, September~2021}%
{Shell \MakeLowercase{\textit{et al.}}: Bare Demo of IEEEtran.cls for Computer Society Journals}
%



\IEEEtitleabstractindextext{%
\begin{abstract}
Fault tree analysis is a well-known technique in reliability engineering and risk assessment, which supports decision-making processes and the management of complex systems. Traditionally, fault tree (FT) models are built manually together with domain experts, considered a time-consuming process prone to human errors. With Industry 4.0, there is an increasing availability of inspection and monitoring data, making techniques that enable knowledge extraction from large data sets relevant. Thus, our goal with this work is to propose a data-driven approach to infer efficient FT structures that achieve a complete representation of the failure mechanisms contained in the failure data set without human intervention. Our algorithm, the FT-MOEA, based on multi-objective evolutionary algorithms, enables the simultaneous optimization of different relevant metrics such as the FT size, the error computed based on the failure data set and the Minimal Cut Sets. Our results show that, for six case studies from the literature, our approach successfully achieved automatic, efficient, and consistent inference of the associated FT models. We also present the results of a parametric analysis that tests our algorithm for different relevant conditions that influence its performance, as well as an overview of the data-driven methods used to automatically infer FT models.
\end{abstract}

}

\maketitle

\IEEEdisplaynontitleabstractindextext

%
\IEEEpeerreviewmaketitle


\IEEEPARstart{F}{ault} Tree Analysis (FTA) is a widely used method in reliability engineering and risk analysis, mainly because it enables modeling complex systems by encoding and displaying logical relationships that can be used, among others, to understand how a system might fail, trace the root cause of the failure, identify critical components, and calculate the system and subsystem failure probabilities.

Fault Tree (FT) models have been around since the 1960s and have been used in a wide range of domains, including the automotive, aerospace, and nuclear industries \cite{kabir2017overview}. However, a major drawback of FTs is related to their construction, which is traditionally carried out in conjunction with domain expertise and in a hand-crafted manner, resulting in a tedious and time-consuming task. In the case of complex industrial systems, manual development of these models can lead to incompleteness, inconsistencies, and even errors \cite{signoret2021automated}.

The above challenge has been discussed since the 1970s, and it is referred to in the literature as \textit{construction} \cite{salem1976computer}, \textit{synthesis} \cite{hunt1993propagation}, or \textit{induction} \cite{madden1970generation} of FTs. In this work, we refer to this as \textit{automatic inference of FT models}. The latter is a process, where a given observational failure data set, automatically outputs the FT structure that best encodes the logic that describes the failure propagation in the system.

Different ways of inferring FTs have been discussed in the past. We identify three main groups namely \textit{knowledge-based}, \textit{model-based}, and \textit{data-driven}. The main differences between these categories are that \textit{knowledge-based} approaches employ different heuristics for knowledge representation, using information about the basic components and their interconnections in the system under analysis \cite{latif2002comparing}; \textit{model-based} approaches translate existing system and/or graph models into FTs, and \textit{data-driven} approaches have structured databases as the primary source of information.

Carpignano \& Poucet \cite{carpignano1994computer} thoroughly review several knowledge-based approaches. An example of a model-based approach can be found in Mhenni et al. \cite{mhenni2014automatic} where the authors used SysML System Models as a base to obtain FT models. However, a major drawback of model-based approaches is the need for a pre-existing model \cite{dickerson2018formal}.

We focus our attention on \textit{data-driven} approaches, where the applications of machine learning techniques and data analytics fall into this category. In Appendix \ref{app:ddFTs_table}, Table \ref{tb:state_of_the_art_literature} (divided into two parts) we summarise and compare relevant literature in data-driven methods for the automatic inference of FT models. Below we briefly discuss these methods.

To the best of our knowledge, the very first attempt to tackle the challenge in a data-driven manner was made by Madden \& Nolan \cite{madden1970generation} with their IFT algorithm, which is based on Quinlan's ID3 algorithm to induce Decision Trees (DTs) \cite{quinlan1986induction}. The authors also continued with this work in the subsequent years \cite{madden1970hierarchically, madden1999monitoring}. Mukherjee \& Chakraborty \cite{mukherjee2007automated} addressed this challenge via \textit{linguistic analysis} and domain knowledge to identify the nature of the failure from short plain text descriptions of equipment faults, and from this generate an FT model. Roth et al. \cite{roth2015integrated} propose a method that follows the Structural Complexity Management (StCM) methodology, where their main goal is to deduce dependencies that are later on are used to infer the Boolean logic operators of the FT models. Inspired by \textit{Causal Decision Trees} \cite{li2016causal}, in Nauta et al. \cite{nauta2018lift} they proposed an approach based on the \textit{Mantel-Haenszel} statistical test.

Based on \textit{Knowledge Discovery in Data set}, Waghen \& Ouali \cite{waghen2019interpretable} propose a method for hierarchical causality analysis called \textit{Interpretable Logic Tree Analysis} (ILTA), that looks for patterns in a data set which are translated into \textit{Interpretable Logic Trees}. Linard et al. \cite{linard2019induction} proposed a method that consists of learning a \textit{Bayesian Network} graph, which is later translated into an FT model. Lazarova-Molnaretal \cite{Feng2020DATADRIVENFT} presents an approach based on time series of failure data, binarization techniques, MCSs, and Boolean algebra. This algorithm also manages to infer FT models with VoT gates. In Waghen \& Ouali \cite{waghen2021multi} the authors further extended their work to the \textit{Multi-level Interpretable Logic Tree Analysis} (MILTA), which tackles the problem of multiple cause-and-effect sequences (the latter a limitation of their previous version, the ILTA algorithm) by incorporating Bayesian probability rules.

The first attempt based on \textit{evolutionary algorithms} was carried out by Linard et al. \cite{linard2019fault}, here the authors created an algorithm to generate FT models from a labeled binary failure data set using a uni-dimensional cost function based on the \textit{accuracy}, which is the proportion of correctly predicted top events by a given FT. 

A drawback of the above approach is that it focuses solely on achieving high accuracy without considering the FT structure. The latter has negative implications such as:  running the algorithm twice for the same failure data set may yield considerably different FT structures; it is prone to complexity explosion, leading to massive FTs that are difficult to handle; long computational time, and bad convergence of the algorithm.

Thus, the main contributions of this paper are that (i) we show it is possible to achieve more consistent and efficient FT structures via \textit{multi-objective evolutionary algorithms} (MOEAs), which  simultaneously optimize various criteria in a multi-dimensional space \cite{deb2014multi}, (ii) we propose a metric to compare FT structures via Minimal Cut Sets using the RV-coefficient, (iii) we carry out a parametric analysis that explains the performance of the algorithm under different assumptions.

The remaining part of this paper is organized as follows. Section \ref{sec:theoretical_background} formally defines FTA and provides the technical background of MOEAs. Section \ref{sec:Methodology} explains our methodology. Section \ref{sec:learning_fts_moga} presents how we apply the NSGA-II (an MOEA) to infer FTs. Section \ref{sec:results} presents the results of a thorough parametric analysis. Section \ref{sec:disucssion_conclusions} discusses our findings and presents our conclusions.

 


\section{Theoretical background}\label{sec:theoretical_background}

\subsection{Fault Tree Analysis}\label{sec:fta}


\textbf{Fault Tree Analysis} (FTA) is one of the most famous methods in reliability engineering that supports decisions in the design and maintenance of complex systems. FTA enables \textit{qualitative} and \textit{quantitative} analyses. Qualitative analysis is based on the FT structure and aims at finding the critical system components. Here, an important concept refers to the \textit{Minimal Cut Sets} (MCSs), which are minimal combinations of component failures that lead to a system failure. Small minimal cut sets point to system vulnerabilities. 


The quantitative analysis aims at computing various dependability metrics,  such as system \textit{Reliability}; \textit{Availability}; and the \textit{Mean-Time-to-Failure}. Estimation of these metrics requires that the FT leaves are equipped with failure probabilities.

A \textbf{Fault Tree} (FT) helps in understanding why a system fails by modeling  how low-level failures propagate through the system and lead to the system-level failure. As similarly done by Ruijters and Stoelinga \cite{ruijters2015fault}, to formally define an FT, we first define $\mathrm{GateType = \{And,Or\} \; \cup }$ $\mathrm{ \{ VoT(k/N) \:  | \:  k, \: N \; \in \: \mathbb{N}^{\geq 1}, \:  k \leq N \}}$. Then, an FT is a 5-tuple $\mathrm{F = \langle BE, G, T, I, \TopEvent \rangle}$ where

\begin{itemize}
	\item $\mathrm{BE}$ is a set of basic events, which may be annotated with a probability of occurrence $\mathrm{p_i}$.
	\item $\mathrm{G}$ is a set of logic gates, with $\mathrm{BE \cap G = \emptyset}$. 
	\item $\mathrm{E=BE \cup G}$ for the set of elements.
	\item $\mathrm{T:G \to GateTypes}$ is a function that describes the type of each gate.
	\item $\mathrm{I:G \to P(E)}$ describes the inputs of each gate. We require $\mathrm{I(g) \neq \emptyset}$ and that $\mathrm{|I(g)| = N}$ if $\mathrm{T(g) = VoT(k/N)}$.
	\item $\mathrm{And}$ gate is a tuple $\mathrm{\langle And, I, O \rangle}$ where $\mathrm{O}$ outputs \textit{true}, if \textit{every} $\mathrm{i \in I}$ occurs.
	\item $\mathrm{Or}$ gate is a tuple $\mathrm{\langle OR, I, O \rangle}$ where $\mathrm{O}$ outputs \textit{true}, if \textit{at least one} $\mathrm{i \in I}$ occurs.
	\item $\mathrm{VoT(k/N)}$ gate is a tuple $\mathrm{\langle VoT, k, I, O \rangle}$ where $\mathrm{O}$ outputs \textit{true}, if \textit{at least} $k$ of $\mathrm{i \in I}$ occur.
	\item $\mathrm{\TopEvent \in E}$ is a unique root called the top event, and it is reachable from all other nodes. 
\end{itemize}

\begin{figure}[!h]
\centering
\includegraphics[width=1\linewidth]{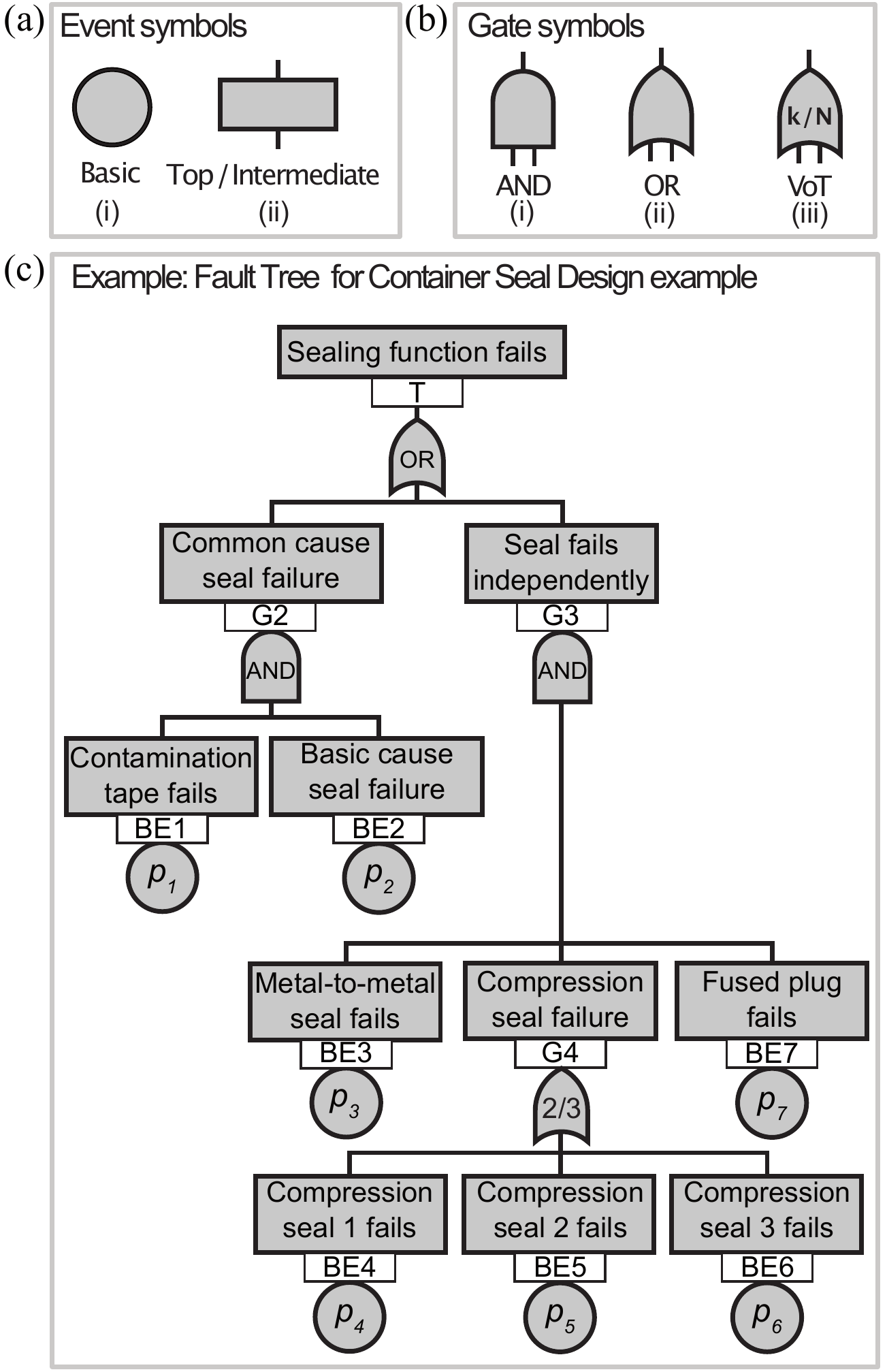}
\caption{Elements in an FT: (a) event symbols, (b) gate symbols, and (c) example of an FT, adapted from \cite{stamatelatos2002fault}.}
\label{fig:gxluLpKpaC_static_fault_tree_structure}
\end{figure}

Importantly, the graph formed by $\mathrm{(E,I)}$ should be a directed acyclic graph. Fig. \ref{fig:gxluLpKpaC_static_fault_tree_structure}.(a) and \ref{fig:gxluLpKpaC_static_fault_tree_structure}.(b) depict respectively the \textit{event} and \textit{gate} symbols used to construct the FT model.

\begin{figure*}[!h]
\centering
\includegraphics[width=1\linewidth]{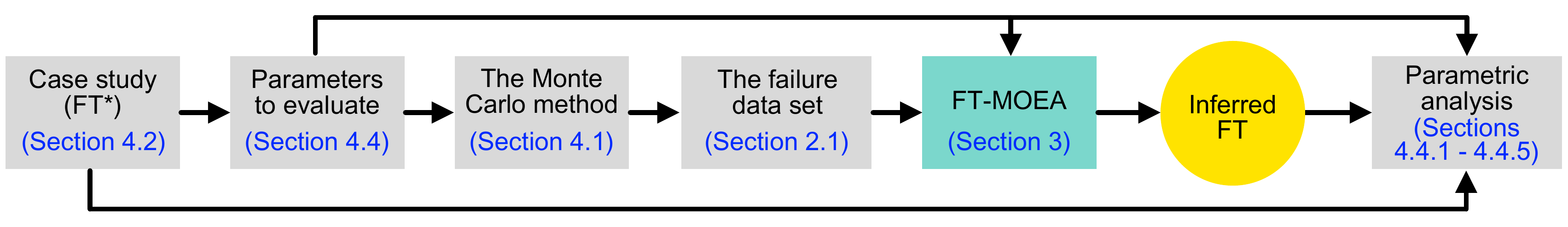}
\caption{General methodology.}
\label{fig:FYrP2nnkB6_general_method}
\end{figure*}

Fig. \ref{fig:gxluLpKpaC_static_fault_tree_structure}.(c) provides an example of an FT. This is the FT of a \textit{Container Seal Design} adapted from \cite{stamatelatos2002fault}. In this FT, the top event, the \textit{sealing function fails}, either occurs if a \textit{common cause seal failure} occurs or if the \textit{seals fail independently}. For the former, it is necessary that the \textit{contamination tape fails}, and a \textit{basic cause seal failure} occurs. For the latter, it is necessary for the \textit{metal-to-metal seal}, the \textit{fused plug}, and at least two of the three \textit{compression seals} to fail. 


\subsection{Multi-Objective Evolutionary algorithms}\label{sec:evolutionary_alg}

\textit{Evolutionary algorithms} (EAs) are population-based search strategies that use the fundamental principle of natural selection, where the best individuals are more likely to reproduce and, therefore, to pass on to the next generations \cite{ojha2019review}. When the EA deals with several conflicting objective functions to be simultaneously optimized in a multi-dimensional space \cite{deb2011multi}, we call these \textit{Multi-Objective Evolutionary Algorithms} (MOEAs). The result of MOEAs is a set of solutions with trade-offs, better known as \textit{Pareto-optimal} solutions, from which a user can then, based on higher-level qualitative considerations, make a choice \cite{deb2014multi}.


As a first step to tackle the challenge of automatically inferring FTs from a failure data set by simultaneously optimizing different metrics, we decided to use one of the most popular algorithms in multi-objective optimization called the \textit{Elitist Non-dominated Sorting Genetic Algorithm} (NSGA-II) (Section \ref{sec:nsgaII}) and the \textit{Crowding-Distance} (Section \ref{sec:crowding_distance}).

\subsubsection{Elitist Non-dominated Sorting Genetic Algorithms (NSGA-II)}\label{sec:nsgaII}

The \textit{Elitist Non-dominated Sorting Genetic Algorithm} (NSGA-II) \cite{deb2002fast} aims at finding multiple Pareto-optimal solutions. NSGA-II uses the \textit{elitist} principle which is a diversity preserving mechanism that emphasizes the non-dominated solutions \cite{deb2014multi}. Moreover, the elitist principle guarantees that the quality of the solution does not decrease, by letting the best individual(s) of the current generation pass to the next one.


Non-dominated MOEAs use the concept of \textit{dominance}, where two solutions are compared to evaluate whether one dominates the other or not. Non-dominated sorting is a crucial step to uncover elitist efficient solutions in MOEAs, but it is computationally expensive since it involves numerous comparisons \cite{long2021non}. A set of solutions that do not dominate each other is known as a \textit{non-dominated front}. In Appendix \ref{sec:app_nsgaII_to_infer_fts} we provide a detailed explanation of these concepts and exemplify how the NSGA-II is applied in the automatic inference of FTs.

\subsubsection{Crowding-Distance}\label{sec:crowding_distance}

Since it may be the case that the last non-dominated front obtained through the NSGA-II does not fully accommodate the available slots to complete the new population, the \textit{Crowding-Distance} is used to decide which individuals of the last front can pass to the next generation, acting as a mechanism that promotes diversity \cite{marti2017impact}. In Appendix \ref{sec:app_crowding_distance} we provide details on how the Crowding-Distance is applied in the automatic inference of FTs.

\section{Methodology}\label{sec:Methodology}

Fig. \ref{fig:FYrP2nnkB6_general_method} depicts the general methodology we followed in this paper. First, we selected some case studies of existing FTs (Section \ref{sec:case_studies}), these FTs act as ground truth. Then, we selected some parameters of interest (Section  \ref{sec:parametric_analysis}) to be evaluated in the parametric analysis. We used the Monte Carlo method (Section \ref{sec:monte_carlo_method}) to generate failure data sets (Section \ref{sec:fault_dataset}) based on the selected case studies. Then we used our FT-MOEA algorithm (Section \ref{sec:learning_fts_moga}) to infer the FT based on the provided failure data set. Finally, we compared the ground truth with the inferred FTs and evaluate the experiment (Section \ref{sec:parametric_pop_size} to \ref{sec:initial_fts_strategies}).


\subsection{The failure data set}\label{sec:fault_dataset}

Our methodology needs the following assumptions for the input failure data set:

\begin{itemize}
	\item \textit{Labeled}: Our data set consists of combinations of $\mathrm{BE}$ and their corresponding $\mathrm{\TopEvent}$.
	\item \textit{Binary}: Both $\mathrm{BE}$ and $\mathrm{\TopEvent}$ are binary. This poses an advantage because we can make use of Boolean operations which makes the algorithm faster. In our case, 0 and 1 are used as non-faulty and faulty states, respectively.
	\item \textit{Monotonic/consistent}: For a given set of $\mathrm{BE}$, if a single $\mathrm{BE}$ changes from state 0 to 1, it is possible that $\mathrm{\TopEvent}$ changes from state 0 to 1, but it will never change from state 1 to 0.
	\item \textit{Complete}: The total number of unique combinations of $\mathrm{BE}$ in the failure data set equals the space complexity $O(2^w)$, where $w$ corresponds to the number of unique $\mathrm{BE}$ for a given FT.
	\item \textit{Noise-free}: There is no corrupted information contained in the failure data set. In other words, the relation $\mathrm{BE \to \TopEvent}$ is always true for a given FT.
\end{itemize}

Table \ref{tb:knzjjUHiLt_example_dataset} presents an example of the data set associated with the FT in Fig. \ref{fig:gxluLpKpaC_static_fault_tree_structure}.  To generate this data set, we used the Monte Carlo method described in Section \ref{sec:monte_carlo_method} for $\mathrm{N=250,000}$ data points and using a failure rate for the $\mathrm{BE}$ of $p_i = 0.5$. \textit{Ob.} refers to the observation associated with a unique combination of values of $\mathrm{BE}$, and the associated $\mathrm{\TopEvent}$. The columns $\mathrm{BE1}$, $\mathrm{BE2}$, ..., $\mathrm{BE7}$ show the states of the set of $\mathrm{BE}$. $\mathrm{\TopEvent}$ corresponds to the top event. Finally, the last column refers to the \textit{count} of each of the observations in the failure data set.

\begin{table}[ht]\setlength\tabcolsep{3pt} 
\centering
\small

\caption{Example of failure data set associated to the example in Fig. \ref{fig:gxluLpKpaC_static_fault_tree_structure}.}
 \begin{tabular}{@{}c|ccccccc|cc@{}}
    \toprule
    \textit{Ob.} & $\mathrm{BE1}$  & $\mathrm{BE2}$ & $\mathrm{BE3}$ & $\mathrm{BE4}$ & $\mathrm{BE5}$ & $\mathrm{BE6}$ & $\mathrm{BE7}$ & $\mathrm{TE}$  & \textit{Count}  \\ \midrule
    1 & 0 & 0 & 0 & 0 & 0 & 0 & 0 & 0 & 1,968\\
    2 & 0 & 0 & 0 & 0 & 0 & 0 & 1 & 0 & 2,039\\
    $\vdots$ & $\vdots$  & $\vdots$ & $\vdots$ & $\vdots$ & $\vdots$ & $\vdots$ & $\vdots$ & $\vdots$ & $\vdots$\\
    24 & 0 & 0 & 1 & 0 & 1 & 1 & 1 & 1 & 1,976\\
    $\vdots$ & $\vdots$  & $\vdots$ & $\vdots$ & $\vdots$ & $\vdots$ & $\vdots$ & $\vdots$ & $\vdots$ & $\vdots$ \\
    128 & 1 & 1 & 1 & 1 & 1 & 1 & 1 & 1 & 1,947\\ \bottomrule
    $p \approx$  & 0.50  & 0.50  & 0.50  & 0.50  &  0.50 &  0.50 & 0.50  & $N =$ & 250,000\\ 
 \end{tabular}\label{tb:knzjjUHiLt_example_dataset}

\end{table}

\section{Inferring fault trees via multi-Objective evolutionary algorithms (FT-MOEA)}\label{sec:learning_fts_moga}

\begin{figure}[!h]
\centering
\includegraphics[width=1\linewidth]{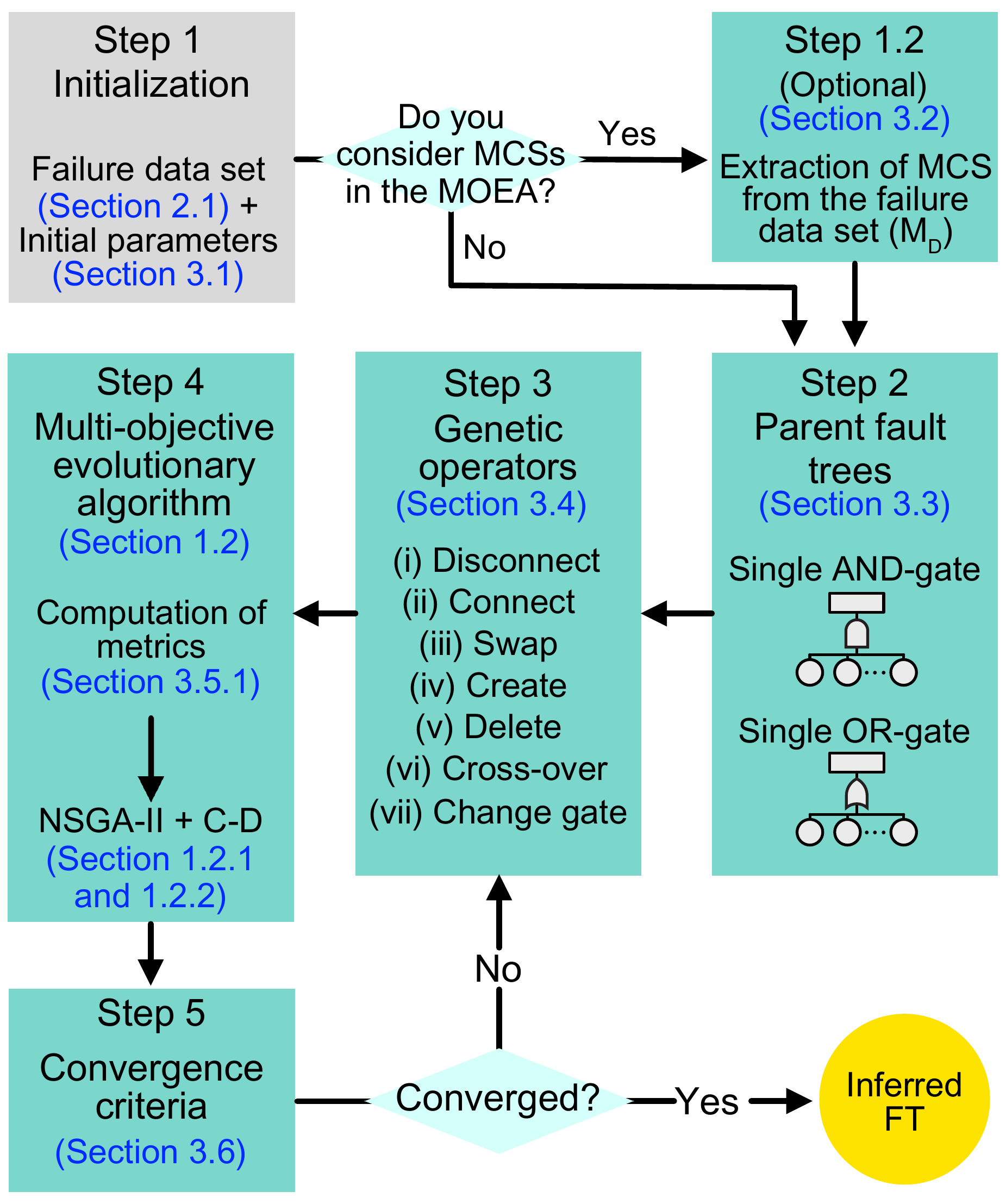}
\caption{General process of the FT-MOEA algorithm to infer FTs from a failure data set.}
\label{fig:mA1tmgvHCn_fig_optimization_process}
\end{figure}

The input of the algorithm is composed of the failure data set, described in Section \ref{sec:fault_dataset}, and the initial parameters of the MOEA, described in Section \ref{sec:initialization}. The output of the algorithm is a string that describes the structure of the inferred FT (based solely on $\mathrm{And}$ and $\mathrm{Or}$ gates), and metrics of interest that tell about the error between the inferred FT and the failure data set. These metrics are explained in detail  in Section \ref{sec:comp_metrics}. Fig. \ref{fig:mA1tmgvHCn_fig_optimization_process} provides an overview of our approach, consisting of the following five steps:

\begin{enumerate}
    \item Step 1: Initialize the algorithm by loading the \textit{failure data set} (Section \ref{sec:fault_dataset}) as well as the initial parameters (Section \ref{sec:initialization}). An optional step is the extraction of the MCSs from the failure data set (Section \ref{sec:extraction_MCSs}), which is executed only when the objective function considers metrics based on MCSs.  
    \item Step 2: Initialize the population with the \textit{parent fault tree(s)} (Section \ref{sec:initial_parent_fts}) on which the genetic operators are applied until the desired population size is reached.
    \item Step 3: Apply the \textit{genetic operators} (Section \ref{sec:genetic_operators}) to randomly modify the structure of the FTs. These genetic operators are recursively applied to a population of FTs until reaching at least the desired population size.
    \item Step 4: For each FT in the offspring population, compute the \textit{metrics} to be optimized (Section \ref{sec:comp_metrics}). Then, using the concept of \textit{Pareto efficiency} through the \textit{Elitist Non-Dominated Sorting Algorithm (NSGA-II)} and \textit{Crowding-Distance} (Section \ref{sec:evolutionary_alg}), determine the population of FTs for the next generation.
    \item Step 5: Check whether any of the convergence criteria (Section \ref{sec:converge_criteria}) are met. If not, the genetic operators (Step 3) are applied to the new population, and Steps 4 and 5 are applied recursively until at least one of the convergence criteria is met, outputting a Pareto set of inferred FTs where we chose the best individual i.e., the FT with the smallest size, and smallest error(s) within the first Pareto set.

\end{enumerate}

\subsection{Step 1 - Initialization}\label{sec:initialization}

We have the following initial parameters.
\begin{itemize}
    \item \textit{Population size} ($ps$): Corresponds to the number of FTs within a generation. Only the best $ps$ FTs can pass to the next generation.
    \item \textit{Selection strategy}: For the NSGA-II algorithm we only use the \textit{elitist} selection strategy.
    \item \textit{Max. generations with unchanged best candidate} ($uc$): if after $uc$ number of generations the best individual (i.e., the FT with the smallest size, and smallest error(s) within the best Pareto set) remains unchanged, then we assume the process has converged and is therefore terminated.  
    \item \textit{Max. number of generations} ($ng$): Terminates the optimization process if the number of generations exceeds $ng$ and none of the other convergence criteria is met. 
\end{itemize}

\subsection{Step 1.2 - Extraction of MCSs from the failure data set (optional step)}\label{sec:extraction_MCSs}

As mentioned before, MCSs are a minimal combinations of component failures that lead to a system failure. This information is extremely valuable because it encodes the failure modes of the system. By considering this information in our optimization process we are adding additional criteria that can help our algorithm to find better solutions in a shorter amount of time. However, the user can decide whether to consider MCSs within the optimization process. 

One should consider MCSs only if it is guaranteed that the failure data set is \textit{noise-free}, and if the expected FT is not too complex (see Section \ref{sec:ft_complexity} where we address the topic of complexity in FTs). The former because otherwise one cannot be sure whether MCSs are correct, and the latter because computing MCSs is computationally expensive (we will discuss this in Section \ref{sec:comp_metrics}) and for complex FTs, this could increase the convergence time.

The process of extracting MCSs from a failure data set is done as described by Lazarova-Molnar et al. \cite{Feng2020DATADRIVENFT} with the following four steps. \textit{Step 1:} identify all the combinations of $\mathrm{BE}$ that output class 1 (i.e., $\mathrm{\TopEvent=1}$). \textit{Step 2:} from this sub-set, identify the one with the minimal order (where the order is defined as the number of \textit{true} $\mathrm{BE}$ in an observation) and save this observation as part of the MCSs matrix computed from the failure data set ($\mathrm{M_{D}}$). \textit{Step 3:} look in the sub-set for other observations that include the previously identified MCS. If any, delete them from the sub-set. \textit{Step 4:} repeat Steps 2 and 3 until the sub-set is empty. 

Now, the obtained $\mathrm{M_{D}}$ can be used as an input argument to compute the accuracy based on the MCSs ($\mathrm{\phi_c}$) (see Eq. \ref{eq:rv}). Table \ref{tb:example_mcs_matrix} provides an example of $\mathrm{M_{D}}$ for the failure data set described in Table \ref{tb:knzjjUHiLt_example_dataset} for the example in Figure \ref{fig:gxluLpKpaC_static_fault_tree_structure}. 

\begin{table}[ht]\setlength\tabcolsep{3pt} 
\centering
\small
\caption{Example of MCS matrix ($\mathrm{M_D}$) computed from the failure data set described in Table \ref{tb:knzjjUHiLt_example_dataset} associated with the example in Figure \ref{fig:gxluLpKpaC_static_fault_tree_structure}.}
 \begin{tabular}{@{}c|ccccccc@{}}
    \toprule
    \textit{MCS} &  \textit{BE1} & \textit{BE2} & \textit{BE3} & \textit{BE4} & \textit{BE5} & \textit{BE6} & \textit{BE7}  \\ \midrule
    1 & 1 & 1 & 0 & 0 & 0 & 0 & 0 \\
    2 & 0 & 0 & 1 & 1 & 1 & 0 & 1 \\
    3 & 0 & 0 & 1 & 0 & 1 & 1 & 1 \\
    4 & 0 & 0 & 1 & 1 & 0 & 1 & 1 \\ \bottomrule
 \end{tabular}\label{tb:example_mcs_matrix}
\end{table}


\subsection{Step 2 - parent fault tree(s)}\label{sec:initial_parent_fts}

The \textit{parent fault tree(s)} can be seen as that one (or those ones) from which the offspring population is generated when applying the \textit{genetic operators} (Section \ref{sec:genetic_operators}). Defining the parent FT(s) is important because it determines how far away, from the global optimum, one starts the optimization process. In Linard et al. \cite{linard2019fault} two parent FTs are used to generate the offspring population, one in which the set of $\mathrm{BE}$ are connected to a single $\mathrm{Or}$ gate, and the other to a single $\mathrm{And}$ gate. We consider this aspect in our parametric analysis, the results are in Section \ref{sec:initial_fts_strategies}.

\subsection{Step 3 - Genetic operators}\label{sec:genetic_operators}

The genetic operators are mathematical operations that seek to modify the structure of an FT. We use the seven genetic operators proposed by Linard et al. \cite{linard2019fault}, which also gives the formal definitions of these operators. For completeness, we here provide a short description.

\textit{(i) G-create}: randomly creates an $\mathrm{And}$ or $\mathrm{Or}$ gate under an existing gate in the set $\mathrm{G}$ for a given FT. \textit{(ii) G-mutate}: randomly selects a gate in the set $\mathrm{G}$ and changes its type (i.e., $\mathrm{Or} \to \mathrm{And}$, or $\mathrm{And} \to \mathrm{Or}$). \textit{(iii) G-delete}: from a given FT takes a gate in the set $\mathrm{G}$ and deletes it, including its children. \textit{(iv) BE-disconnect}: from a given FT takes a basic event in the set $\mathrm{BE}$ and disconnects it. 	 \textit{(v) BE-connect}: from a given FT takes a disconnected basic event and randomly places it under a gate in the set $\mathrm{G}$. \textit{(vi) BE-swap}: from a given FT takes a basic event in the set $\mathrm{BE}$ and randomly moves it under a different parent gate in the set $\mathrm{G}$. \textit{(vii) Crossover}: randomly chooses two FTs in the offspring population, then an element in the set $\mathrm{E}$ of each FT is randomly selected and exchanged.

\subsection{Step 4 - Multi-objective function}


In Section \ref{sec:comp_metrics} we define the metrics to be minimized, and in Section \ref{sec:setups_mofs} we describe the different setups of our multi-objective optimization function.

\subsubsection{Computation of metrics}\label{sec:comp_metrics}

In this multi-objective function, we consider three metrics, namely the \textit{fault tree size} ($\FTSize$), the \textit{error based on the failure data set} ($\ErrorData$), and the \textit{error based on the MCSs} ($\ErrorMCSs$).

\begin{itemize}
    \item \textit{Fault Tree Size} ($\FTSize$): corresponds to the number of elements in the FT i.e., the value of $\mathrm{E}$ in the FT (Eq. \ref{sec:eq_tree_size}).
    
    \begin{equation}
    \phi_{s} = |\mathrm{E}| = |\mathrm{BE}| + |\mathrm{G}|  \label{sec:eq_tree_size}
    \end{equation}
    
    Here note that $\FTSize \geq 2$, as each FT has at least one $\mathrm{BE}$ and one $\mathrm{G}$ (i.e., the $\mathrm{\TopEvent}$).

    \item \textit{Error based on the failure data set} ($\ErrorData$): 
    First we need to compute a one-dimensional vector $\mathrm{P}$ with $\mathrm{N}$ values, where $\mathrm{N}$ is the number of data points. $\mathrm{P}$ contains the value of the $\mathrm{\TopEvent}$ of an FT for a given set $\mathrm{BE}$. For the same set $\mathrm{BE}$, we count with the value of the ground truth top event ($\mathrm{TE^*}$), which is in the failure data set (Section \ref{sec:fault_dataset}). Thus, $\ErrorData$ is computed as follows 
    
    \begin{equation}
    \ErrorData = 1 - \frac{\sum_{i=1}^\mathrm{N} x_i}{\mathrm{N}} \left \{ \begin{matrix} x_i = 1 \text{, \textit{if} $\mathrm{P_i = TE^ *_i}$.}
    \\ x_i = 0 \text{, Otherwise.}\end{matrix}\right.\label{eq:acc}
    \end{equation}

    Here $\ErrorData$ varies in the closed interval $[0,1]$, where 0 means that an FT manages to perfectly map the given set $\mathrm{BE}$ in the failure data set into the corresponding $\mathrm{\TopEvent}$ in the failure data set.

    
    \item \textit{Error based on the MCSs} ($\ErrorMCSs$): we propose to compute the error of an FT based on MCSs by means of the \textit{RV-coefficient} \cite{robert1976unifying}. The RV-coefficient is a generalization of the \textit{squared Pearson correlation coefficient} and measures the similarity between the MCS matrix computed from the failure data set ($\mathrm{M_D}$) (see Section \ref{sec:extraction_MCSs}) and the MCS matrix of a given FT ($\mathrm{M_F}$). 
    
    $\mathrm{M_F}$ is computed based on the \textit{disjunctive normal form} (DNF), which in Boolean logic is also known as an $\mathrm{Or}$ of $\mathrm{And}$s. Thus, we transform a given FT into its DNF and identify from it the MCSs to construct $\mathrm{M_F}$. However, this transformation showed to be computationally expensive for large FT sizes. 
    
    In our context, $\mathrm{M_{D}}$ is a $p \times w$ matrix, $\mathrm{M_{F}}$ is a $q \times w$ matrix, $w$ corresponds to the number of unique $\mathrm{BE}$s considered in the problem and $p$ and $q$ are the number of MCSs in the failure data set and an FT, respectively. The computation of $\ErrorMCSs$ is defined as follows
    
    \begin{equation}
    \ErrorMCSs = \mathrm{1 -  \frac{tr(M_{D}M_{F}^TM_{F}M_{D}^T)}{\sqrt{tr(M_{D}M_{D}^T)^2tr(M_{F}M_{F}^T)^2}}}\label{eq:rv}
    \end{equation}
    
    Here $tr(.)$ is the \textit{trace}, and $\ErrorMCSs$ varies in the closed interval $[0,1]$, where $0$ indicates perfect correlation or similarity between $\mathrm{M_{D}}$ and $\mathrm{M_{F}}$. We choose to use the RV-coefficient as a means to compute the error based on MCSs because, for a given problem, the number of unique $\mathrm{BE}$s always remains the same, but the FTs within a population for a given generation often have different number of MCSs with respect the ones found in the failure data set (i.e., $p\neq q$).

\end{itemize}

\subsubsection{Setups of the multi-objective functions}\label{sec:setups_mofs}


Since our multi-objective function (m.o.f.) has three arguments, we can play with different setups and assess their influence (see Section \ref{sec:objective_functions} for the results of the parametric analysis). For example, if we want to minimize only the error based on the MCSs ($\ErrorMCSs$), we can ``turn off'' $\FTSize$ and $\ErrorData$ by assigning constant values (i.e., $\phi_{s} = \phi_{d} = 1$). To differentiate between the different configurations in the m.o.f., we propose the nomenclature presented in Table \ref{tb:setups_mof}. Here `x' refers to whether a metric is being considered (or active) in the m.o.f.

\begin{table}[ht]
\centering
\caption{Different setups of the m.o.f.}
 \begin{tabular}{@{}c|ccc@{}}
    \toprule
    \textit{m.o.f.} &  $\FTSize$ &  $\ErrorData$ &  $\ErrorMCSs$  \\ \midrule
    \textit{sdc} & x & x & x \\
    \textit{dc}  &   & x & x \\
    \textit{sc}  & x &   & x \\
    \textit{sd}  & x & x &   \\
    \textit{c}   &   &   & x \\
    \textit{d}   &   & x &   \\
 \end{tabular}\label{tb:setups_mof}
\end{table}

\subsection{Step 5 - Convergence criterion}\label{sec:converge_criteria}

Our convergence criterion is based on two initial parameters namely the \textit{max. number of generations} ($ng$), and the \textit{max.\ generations with unchanged best candidate} (see Section \ref{sec:initialization}). Additionally, we terminate the convergence process if $\phi_c = 0$ or $\phi_d= 0$ when the minimization of the FT size is ``turned off'', i.e. for the m.o.f.'s $cd$, $c$, or $d$.
\section{Experimental evaluation}\label{sec:results}

For our experimental evaluation, we first selected six case studies from the literature (Section \ref{sec:case_studies}), then we made the implementation of our FT-MOEA algorithm in Python\footnote{https://gitlab.utwente.nl/jimenezroala/ft-moea}. We evaluate our algorithm using synthetic failure data sets (Section \ref{sec:monte_carlo_method}). Some key findings and our parametric analysis are presented respectively in Sections \ref{sec:multi_inside} and \ref{sec:parametric_analysis}.

\subsection{The Monte Carlo method}\label{sec:monte_carlo_method}

We use the Monte Carlo method to generate synthetic failure data sets using the case studies presented in Section \ref{sec:case_studies} and keeping the same properties of the input data set described in Section \ref{sec:fault_dataset}. To generate the synthetic dataset (i) we randomly generate (\textit{N}) data points, by drawing the $\mathrm{BE}$ \textit{independently} from a \textit{binomial distribution} with a probability of success equal to $p_i$, where $i$ corresponds to a basic event, and (ii) computing the corresponding $\mathrm{TE}$ by following the logical rules of the given FT. As one of the main requirements for the failure data set is to be \textit{complete} (see Section \ref{sec:fault_dataset}) we satisfied this condition for each case study by drawing enough data points from the Monte Carlo simulation ensuring that the number of unique observations of $\mathrm{BE}$ equals the space complexity $O(2^w)$, where $w$ corresponds to the number of unique $\mathrm{BE}$ for a give FT.

\subsection{Case studies}\label{sec:case_studies}

In order to establish a sensible ground truth, we used existing FTs in the literature with different applications. Our selection criteria were the number of elements in the FT as well as the number of MCSs and their orders. Table \ref{tb:case_studies} presents the case studies we selected together information on their number of $\mathrm{BE}$, the total number of $\mathrm{And}$, $\mathrm{Or}$, and $\mathrm{VoT}$ gates, the number of MCSs, their orders, and the space complexity which is measured as $O(2^w)$. We make a distinction between the number of unique $\mathrm{BE}$ ($w$) and the total number of $\mathrm{BE}$ ($W$) because some FTs have shared $\mathrm{BE}$.

\begin{table*}[ht]
\centering
\caption{Case studies and associated relevant information. The number of: unique $\mathrm{BE}$s ($w$), total $\mathrm{BE}$s ($W$), $\mathrm{AND}$ gates (\#AND), $\mathrm{OR}$ gates (\#OR), $\mathrm{VoT}$ gates (\#VoT), Minimal Cut Sets (\#MCS), order of MCSs (O-MCSs), and space complexity ($O(2^w)$). Case studies: Container Seal Design (\textbf{CSD}); Pressure Tank (\textbf{PT}); COVID-19 infection risk (\textbf{COVID-19}); Data-driven Fault Tree (\textbf{ddFT}); Mono-propellant propulsion system (\textbf{MPPS}); Spread Monitoring System (\textbf{SMS}).}
 \begin{tabular}{@{}lllllllllll@{}}
    \toprule
    \textit{Case study} & \textit{Ref.} & $w$ & $W$ & \textit{\#AND} & \textit{\#OR} & \textit{\#VoT} & $FT_{size}$ & \textit{\#MCSs}  & \textit{O-MCSs} & \textit{$O(2^w)$}  \\ \midrule
    CSD      &  \cite{stamatelatos2002fault} & 6  & 6  & 2 & 2 & 0 & 10 & 3  & \{2,3,3\} & 64 \\
    PT       &  \cite{stamatelatos2002fault} & 6  & 6  & 1 & 4 & 0 & 11 & 5  & \{1,1,2,2,2\} & 64 \\
    COVID-19 &  \cite{bakeli2020covid} & 9  & 21 & 9 & 3 & 0 & 33 & 6  & \{3,4,4,4,4,4\} & 512  \\
    ddFT     &  \cite{Feng2020DATADRIVENFT} & 8  & 8  & 3 & 1 & 1 & 13 & 6  & \{3,4,4,5,6,6\} & 256  \\
    MPPS     &  \cite{stamatelatos2002fault} & 8  & 12 & 3 & 8 & 0 & 23 & 7  & \{2,2,2,2,2,2,2\} & 256  \\
    SMS      &  \cite{mentes2011application} & 13 & 17 & 0 & 8 & 0 & 25 & 13 & \{1,1,1,1,1,1,1,1,1,1,1,1,1\} & 8,192  \\ \bottomrule
 \end{tabular}\label{tb:case_studies}
\end{table*}

\subsection{Key findings of the FT-MOEA algorithm}\label{sec:multi_inside}

To illustrate our findings and main contributions, we will use the case study Mono-propellant propulsion system (MPPS) from Table \ref{tb:case_studies}. We first generate the failure data set as described in Section \ref{sec:monte_carlo_method}, with $N=250.000$ data points. Then, we used this failure data set as part of the input data of the FT-MOEA algorithm, together with the following initial parameters: $ps=400$, $ng=100$, and $uc=20$.

We first compare the evolutionary process between generations for two m.o.f.'s $d$ and $sdc$. In this way we can respectively compare the approach by Linard et al. \cite{linard2019fault} (only minimizing $\ErrorData$) and our multi-objective optimization process (minimizing $\FTSize$, $\ErrorData$, and $\ErrorMCSs$).

%

\begin{figure}[!h]
\centering
\begin{subfigure}{.24\textwidth}
  \centering
  \includegraphics[width=1\linewidth]{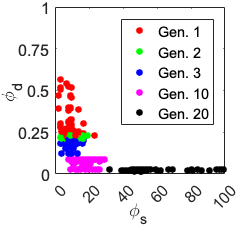}
  \caption{}
\end{subfigure}
\begin{subfigure}{.24\textwidth}
  \centering
  \includegraphics[width=1\linewidth]{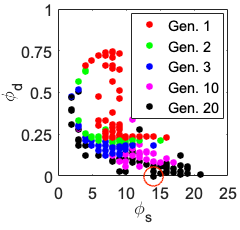}
  \caption{}
\end{subfigure}
\caption{Evolution of metrics over generations. In (a) using the m.o.f. $d$, in (b) using the m.o.f. $sdc$, for the MPPS case study ($ps=400$, $ng=100$, $uc=20$). The red circle at the bottom of (b) indicates the global optimum.}
\label{fig:PyH5PRA7Qw_fig_convergence_between_few_generations}
\end{figure}

 Fig. \ref{fig:PyH5PRA7Qw_fig_convergence_between_few_generations}.(a) shows the results of different generations minimizing solely $\ErrorData$. One can observe that within the first generations there is a rapid decrease in $\ErrorData$ but from the 10th generation onwards, there is a rapid growth in the size of the FTs ($\FTSize$) (up to $\FTSize=100$) without decreasing $\ErrorData$, whereas the ground truth FT is $\FTSize=23$. On the other hand, by using the FT-MOEA algorithm (Fig. \ref{fig:PyH5PRA7Qw_fig_convergence_between_few_generations}.(b)) we observe a smoother decrease in all directions. Additionally, we observe that the FT-MOEA found the global optimum (i.e., $\ErrorData = \ErrorMCSs = 0.0$) in the 20th generation with $\FTSize = 14$ i.e., a compressed version of the ground truth.

\begin{figure}[!h]
\centering
\begin{subfigure}{.22\textwidth}
  \centering
  \includegraphics[width=1\linewidth]{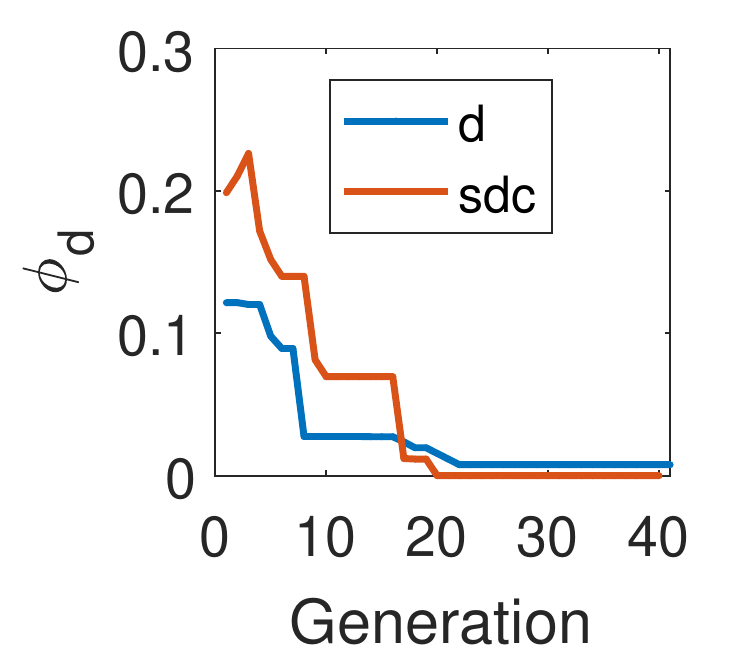}
  \caption{}
\end{subfigure}%
\begin{subfigure}{.22\textwidth}
  \centering
  \includegraphics[width=1\linewidth]{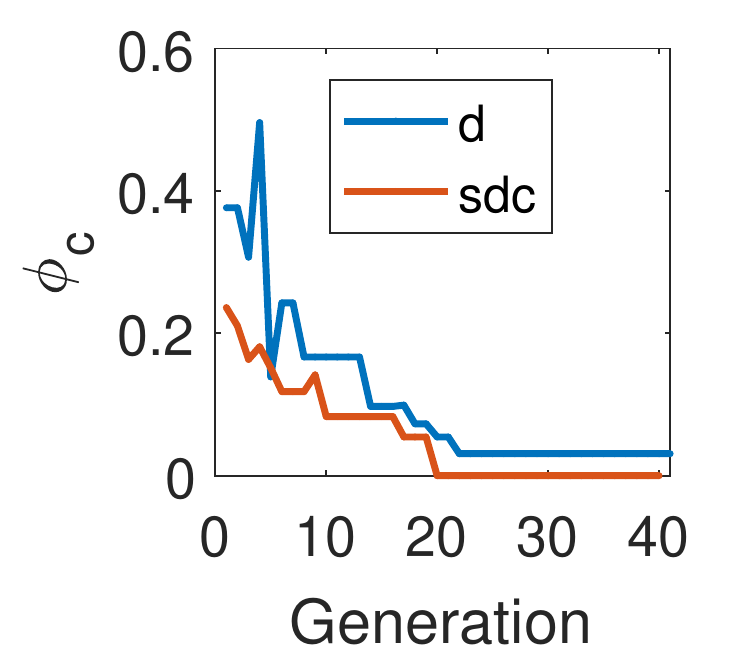}
  \caption{}
\end{subfigure}

\begin{subfigure}{.22\textwidth}
  \centering
    \includegraphics[width=1\linewidth]{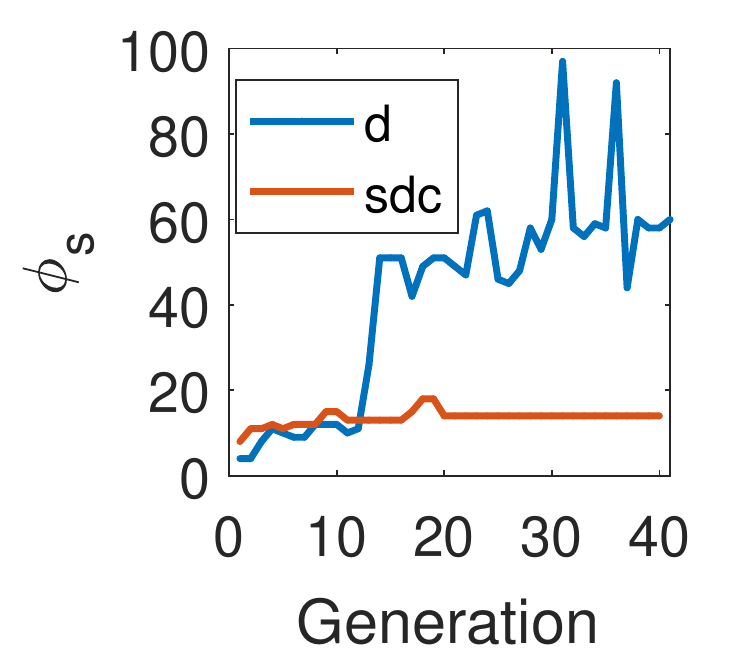}
  \caption{}
\end{subfigure}
\begin{subfigure}{.22\textwidth}
  \centering
  \includegraphics[width=1\linewidth]{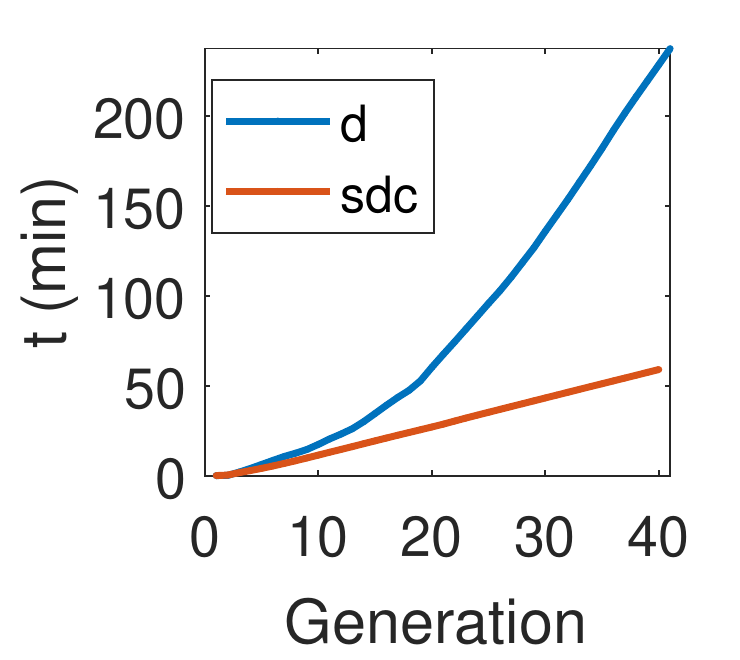}
  \caption{}
\end{subfigure}


\caption{Metrics over generations for the best FTs using the m.o.f.'s $d$ and $sdc$. Convergence of (a) $\ErrorData$, (b) $\ErrorMCSs$, (c) $\FTSize$, (d) cumulative time to convergence ($t$). Using the MPPS case study ($ps=400$, $ng=100$, $uc=20$).}
\label{fig:MmDSvSeoKf_output_ft_comparison_over_generations}
\end{figure}

In Fig. \ref{fig:MmDSvSeoKf_output_ft_comparison_over_generations} we compare both m.o.f.'s across generations using only the metrics of the best FT per generation (i.e., that one in the first Pareto front that has the smallest error $\ErrorData$ and $\ErrorMCSs$). In Fig. \ref{fig:MmDSvSeoKf_output_ft_comparison_over_generations}.(a) we analyze $\ErrorData$ across the generations for both objective functions, we can see that the m.o.f. \textit{d} more rapidly minimize $\ErrorData$ compared to the m.o.f. \textit{sdc}. However, the latter m.o.f. managed to achieve the global optimum in the 20th generation, whereas the former m.o.f. did not find the global optimum. Fig. \ref{fig:MmDSvSeoKf_output_ft_comparison_over_generations}.(b) compares $\ErrorMCSs$, with pretty similar results.

 Fig. \ref{fig:MmDSvSeoKf_output_ft_comparison_over_generations}.(c) depicts the variation of $\FTSize$ over the generations. One can observe that our m.o.f. keeps smaller FT structures, and although the size of the ground truth FT is 23, the FT-MOEA found one of $\FTSize = 14$, i.e., a compressed version of the original one (for more details see Appendix \ref{app:example_output_fts}, Fig. \ref{fig:BF7w7dVZI1_MPPS_fault_tree})
 
 Fig. \ref{fig:MmDSvSeoKf_output_ft_comparison_over_generations}.(d) depicts the cumulative time to convergence ($t$) for both m.o.f.'s. We can observe that our algorithm manages to find the optimal solution in about 20 minutes. On the other hand, by just minimizing $\ErrorData$ the process takes about 4 hours. In Appendix \ref{app:convergence_over_generations} we provide details on the convergence of metrics over the generations for the whole population.

\subsection{Parametric analysis}\label{sec:parametric_analysis}


We consider five aspects to evaluate within our parametric analysis namely the population size, the multi-objective functions, the FT complexity, the influence of superfluous variables, and the initial parent FTs. We decided to explore these parameters because we believe they have an important impact on the computational time, convergence, and the chances of reaching the global optimum.

For this parametric analysis, we generate the failure data set as described in Section \ref{sec:monte_carlo_method}. Since the evolutionary algorithm is a stochastic process, we run our algorithm five times per combination of parameters, and by using \href{https://www.mathworks.com/help/matlab/ref/boxchart.html}{\textit{box charts}} in Matlab (e.g., Fig. \ref{fig:d9i4sRrSGV_parametric_pop_size}) we depict the groups of numerical data through their quartiles.

\subsubsection{Population size}\label{sec:parametric_pop_size}

\begin{figure}[!h]
\centering
\begin{subfigure}{.22\textwidth}
  \centering
  \includegraphics[width=1\linewidth]{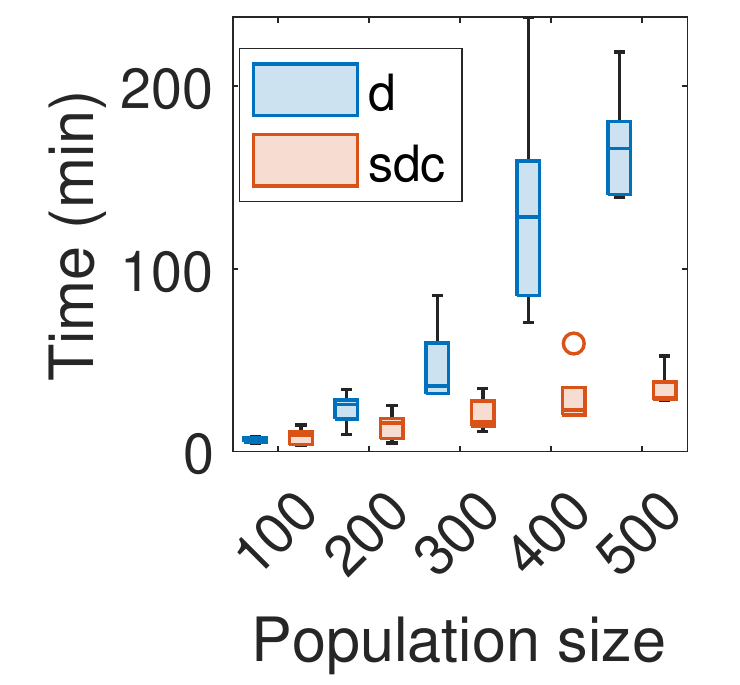}
  \caption{}
\end{subfigure}
\begin{subfigure}{.22\textwidth}
  \centering
  \includegraphics[width=1\linewidth]{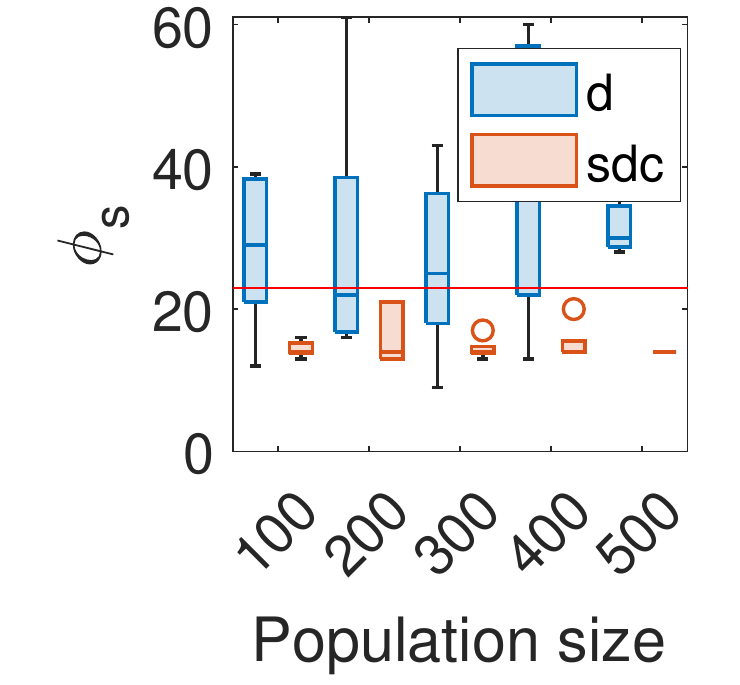}
  \caption{}
\end{subfigure}

\begin{subfigure}{.22\textwidth}
  \centering
  \includegraphics[width=1\linewidth]{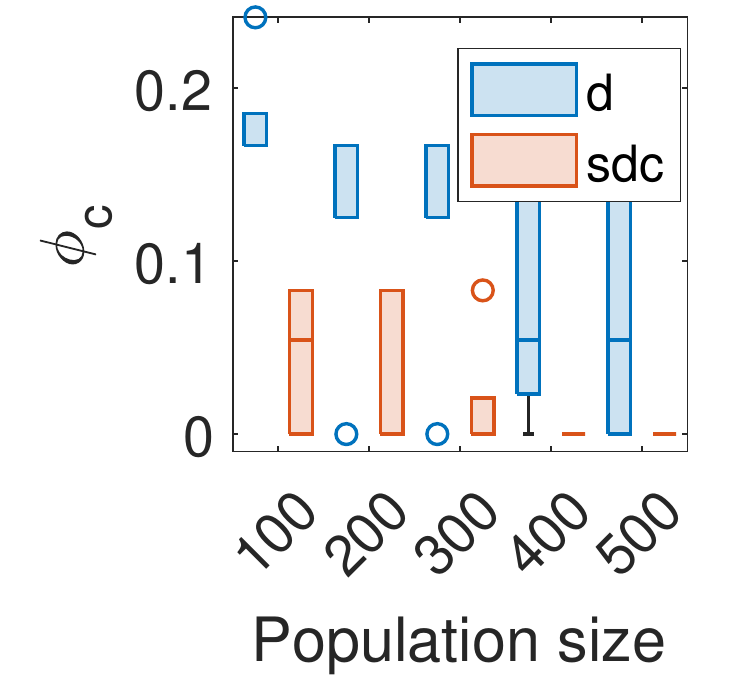}
  \caption{}
\end{subfigure}%
\begin{subfigure}{.22\textwidth}
  \centering
  \includegraphics[width=1\linewidth]{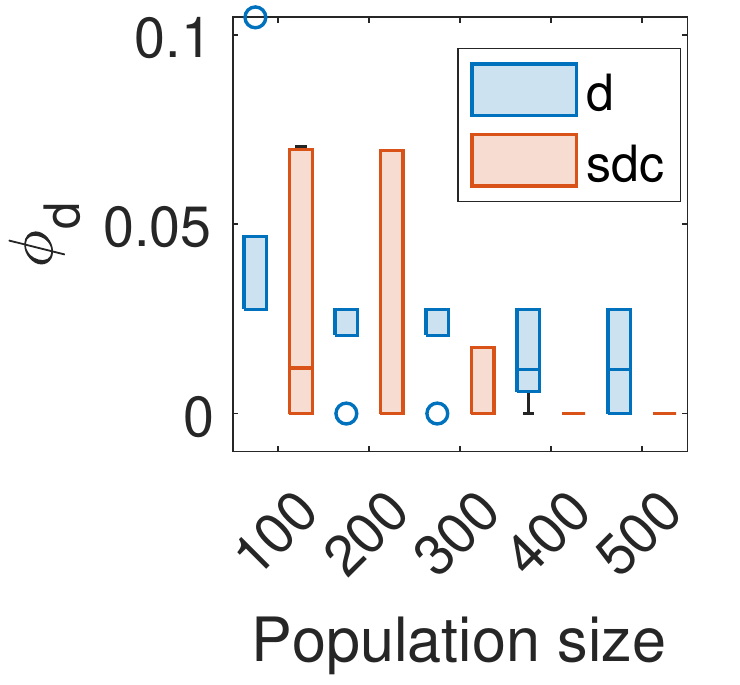}
  \caption{}
\end{subfigure}%
\caption{Influence of population size ($ps$) on (a) convergence time, (b) $\FTSize$, (c) $\ErrorMCSs$, (d) $\ErrorData$. For the m.o.f's $sdc$ and $d$, for the case study MPPS ($ps=400$, $ng=100$, $uc=20$).}
\label{fig:d9i4sRrSGV_parametric_pop_size}
\end{figure}

Fig. \ref{fig:d9i4sRrSGV_parametric_pop_size} presents the results of the parametric analysis when varying the population size for the m.o.f.'s $d$ and $sdc$. Here we use the MPPS case study (Table \ref{tb:case_studies}).

Fig. \ref{fig:d9i4sRrSGV_parametric_pop_size}.(a) shows that larger population sizes exponentially increases the computational time for both m.o.f.'s. However, the m.o.f. $sdc$ is consistently faster. Fig. \ref{fig:d9i4sRrSGV_parametric_pop_size}.(b) shows that the m.o.f. $sdc$ always retrieves smaller FTs compared with the m.o.f. $d$, even smaller than the ground truth (i.e., $\FTSize \leq 23$) indicated by the horizontal red line. From Fig. \ref{fig:d9i4sRrSGV_parametric_pop_size}.(c) and \ref{fig:d9i4sRrSGV_parametric_pop_size}.(d), the m.o.f. \textit{sdc} is more consistent when the population size is larger, also both errors ($\ErrorMCSs$ and $\ErrorData$) tend to decrease with larger population sizes. On the other hand, when using the m.o.f. \textit{d}, it seems that the errors decrease for larger population sizes, but with less consistency. 

\subsubsection{Multi-objective functions}\label{sec:objective_functions}

\begin{figure}[!h]
\centering
\begin{subfigure}{.23\textwidth}
  \centering
  \includegraphics[width=1\linewidth]{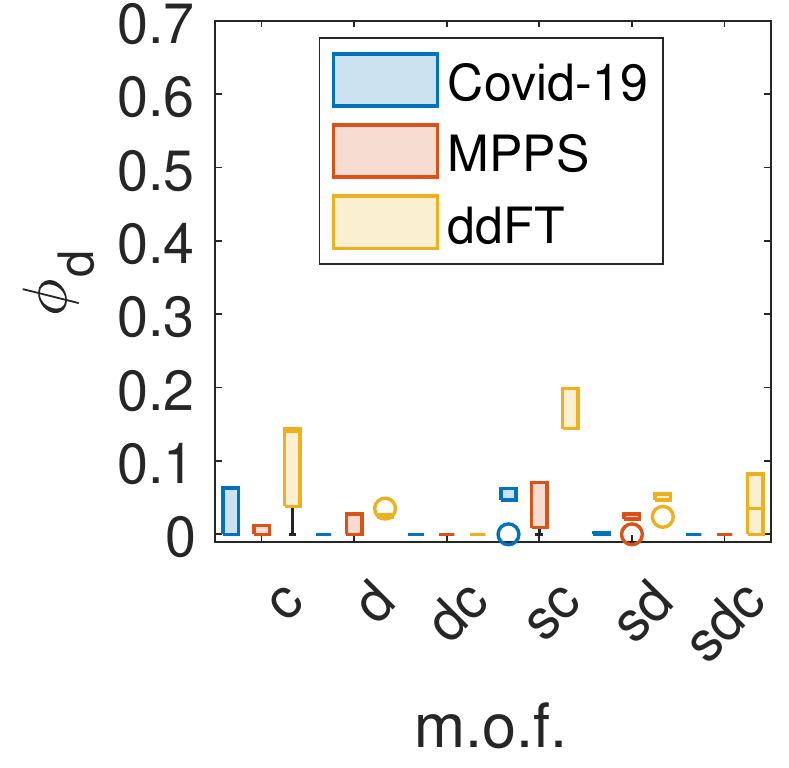}
  \caption{}
\end{subfigure}
\begin{subfigure}{.23\textwidth}
  \centering
  \includegraphics[width=1\linewidth]{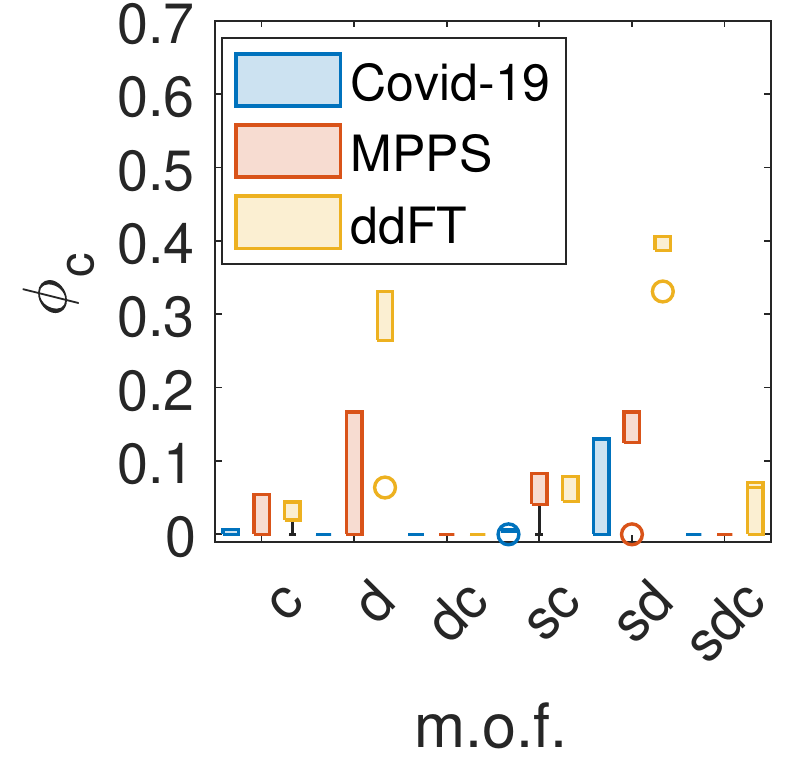}
  \caption{}
\end{subfigure}

\begin{subfigure}{.23\textwidth}
  \centering
  \includegraphics[width=1\linewidth]{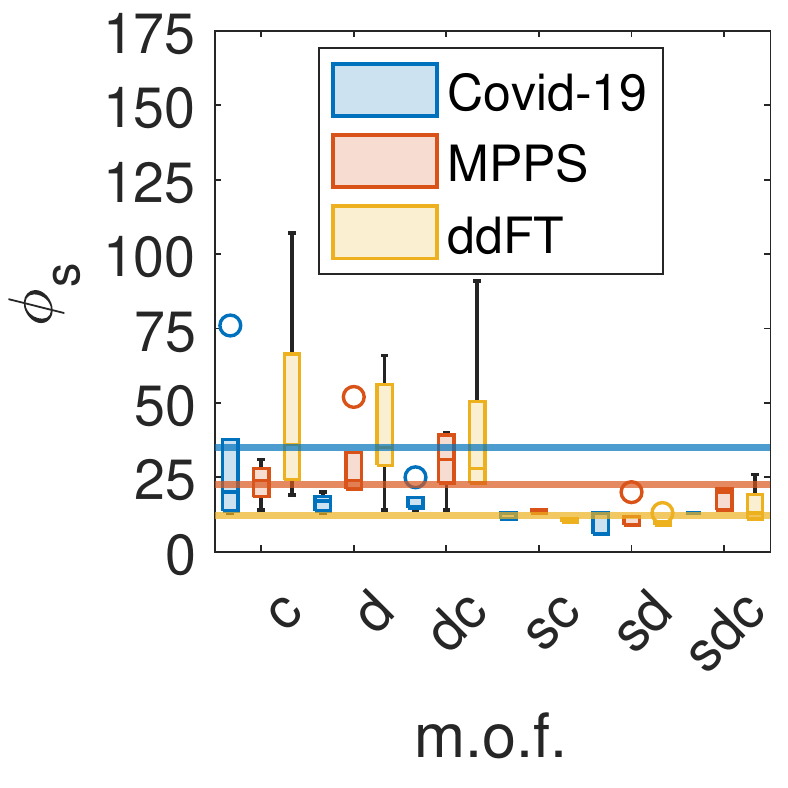}
  \caption{}
\end{subfigure}%
\begin{subfigure}{.23\textwidth}
  \centering
  \includegraphics[width=1\linewidth]{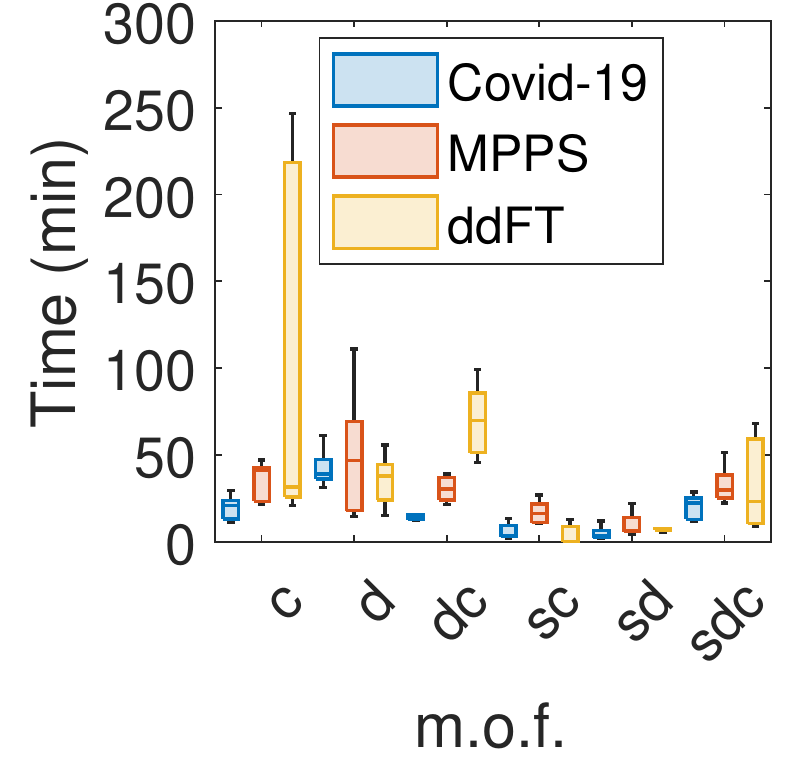}
  \caption{}
\end{subfigure}%
\caption{Comparing the performance of m.o.f.'s, using $ps=400$, $ng=100$, $uc=20$ and the case studies COVID-19, MPPS, and ddFT. In (a) $\ErrorData$, (b) $\ErrorMCSs$, (c) $\FTSize$, (d) convergence time.}
\label{fig:KCaYZmwRj4_comparing_objective_functions}
\end{figure}

We evaluate all Setups of our m.o.f. (Table \ref{tb:setups_mof}), for this, we use the case studies in Table \ref{tb:case_studies}, and keep fixed the input parameters $ps=400$, $ng=100$, $uc=20$. Fig. \ref{fig:KCaYZmwRj4_comparing_objective_functions} presents the results for the case studies COVID-19, MPPS, and ddFT. Fig. \ref{fig:KCaYZmwRj4_comparing_objective_functions2} (Appendix \ref{app:comp_performance_cases_CSD_PT_SMS}) presents the results for the case studies CSD, PT, and SMS.

Fig. \ref{fig:KCaYZmwRj4_comparing_objective_functions}.(a) shows the error based on the failure data set ($\ErrorData$). We observe different behaviors per objective function. The m.o.f. \textit{dc} achieves the exact solution for all the cases, whereas the other m.o.f.'s failed in at least one of the cases to find the global optimum.


Similarly, Fig. \ref{fig:KCaYZmwRj4_comparing_objective_functions}.(b) shows the error based on MCSs ($\ErrorMCSs$). As expected, we observe that the m.o.f. \textit{dc} achieves $\ErrorMCSs = 0.0$ for all the cases. Nevertheless, notice that a low $\ErrorData$ does not imply an optimal FT, as an example, compare $\ErrorData$ and $\ErrorMCSs$ for the case MPPS for the m.o.f.'s $d$ and $sd$.

Fig. \ref{fig:KCaYZmwRj4_comparing_objective_functions}.(c) shows the sizes of the inferred FTs ($\FTSize$). Here we can clearly observe the influence of minimizing $\FTSize$. When accounted, $\FTSize$ for most of the cases is equal or less than the ground truth, indicated with the horizontal lines for the different case studies (see Appendix \ref{app:example_output_fts} for examples and details). When not accounted for, in some cases, $\FTSize$ may be significantly larger than the ground truth.
 


\subsubsection{Fault tree complexity}\label{sec:ft_complexity}

Fig. \ref{fig:KCaYZmwRj4_comparing_objective_functions}.(d) depicts the convergence time. We observe that in general, for all the m.o.f.'s, sorting the case studies based on the convergence time from the longest to the shortest, we have \textit{ddFT}, \textit{MPPS}, \textit{COVID-19}, \textit{CSD}, \textit{PT}, and \textit{SMS}. This indicates that there is a relationship between the \textit{complexity} of the underlying FT model of a failure data set and the time that takes the algorithm to find it.

We believe this complexity is ruled by the amount of MCSs and their orders (see \textit{O-MCSs} in Table \ref{tb:case_studies}). For example, the case study \textit{ddFT} has six MCSs and with orders between 3 and 6, the FT-MOEA generally took the longest time to converge. In contrast, the \textit{SMS} case study has 13 MCSs but all have order 1, here the FT-MOEA converged almost immediately. Thus, the greater the number of MCSs and their orders, the longer will take the algorithm to retrieve the global optimum. We believe that further research is needed to better quantify this type of complexity.




\subsubsection{Influence of superfluous variables}

\begin{figure}[!h]
\centering
\begin{subfigure}{.24\textwidth}
  \centering
  \includegraphics[width=1\linewidth]{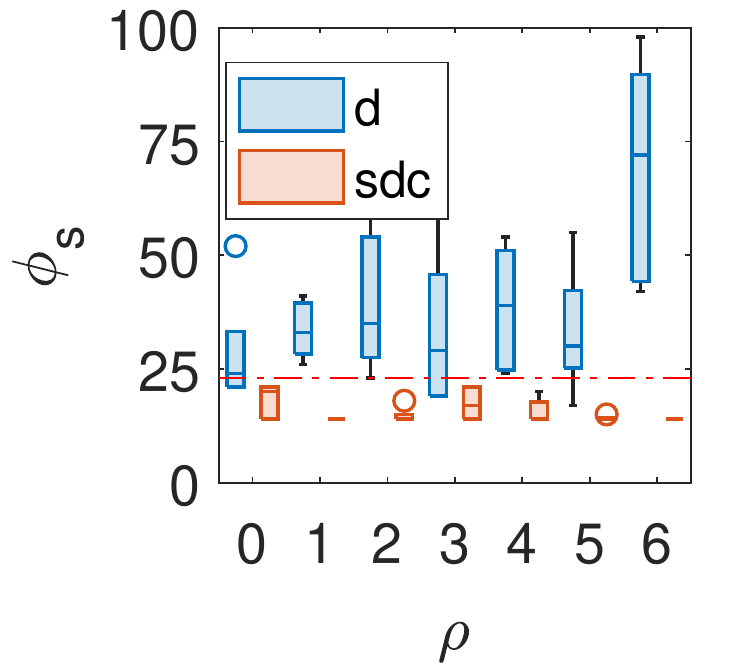}
  \caption{}
\end{subfigure}
\begin{subfigure}{.24\textwidth}
  \centering
  \includegraphics[width=1\linewidth]{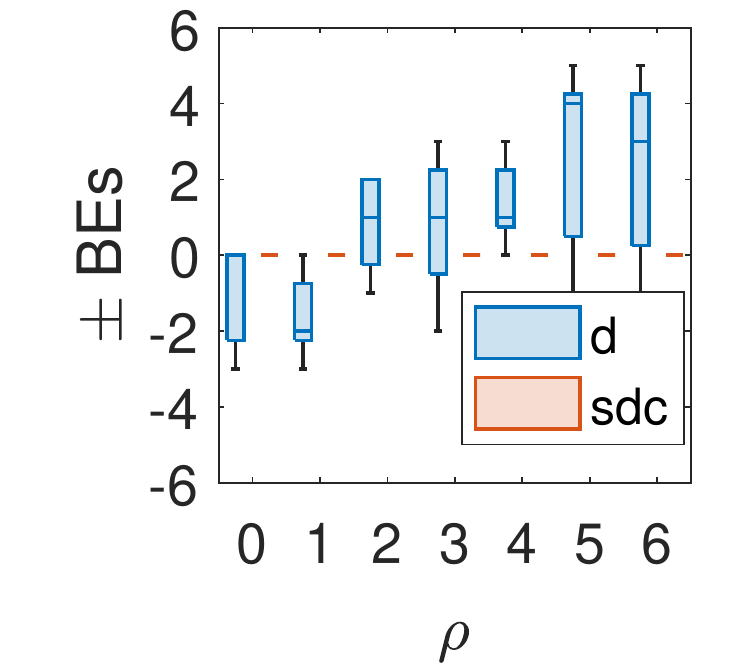}
  \caption{}
\end{subfigure}
\caption{Influence of \textit{superfluous variables} ($\rho$) on (a) size of the resulting FT ($\FTSize$), (b) additional or missing number of $BE$s ($\pm$ $BE$s). Using the m.o.f's $sdc$ and $d$, and the MPPS case study ($ps=400$, $ng=100$, $uc=20$).}
\label{fig:DU6p6C2ci0_spurious_variables}
\end{figure}

Real-world data sets may consider a different number of $\mathrm{BE}$s where possibly not all of them contribute to the failure of the system. In other words, regardless of the state of superfluous $\mathrm{BE}$s, they will not have any effect on the $\mathrm{TE}$. We call these \textit{superfluous variables} ($\rho$). We assess $\rho$ ranging from 0 to 6 using the MPPS case study with $ps=400$, $ng=100$, and $uc=20$, and the m.o.f.'s $sdc$ and $d$. The results are presented in Fig \ref{fig:DU6p6C2ci0_spurious_variables}. 

%
%
%

Fig. \ref{fig:DU6p6C2ci0_spurious_variables}.(a) presents $\FTSize$ for different values of $\rho$. Here $\FTSize$ is smaller than the ground truth (indicated with the dashed horizontal red line) when using the m.o.f. $sdc$. On the other hand, when using the m.o.f. $d$, $\FTSize$ seems to increase for a larger number of $\rho$.


Fig. \ref{fig:DU6p6C2ci0_spurious_variables}.(b) shows for different values of $\rho$ the additional or missing number of $\mathrm{BE}$s ($\pm$ $\mathrm{BE}$s). Recall that the case study MPPS has 7 unique $\mathrm{BE}$s. Thus, we subtract this number from the unique number of $\mathrm{BE}$s per each inferred FT. As we can observe, the m.o.f. $sdc$, despite $\rho$, it always outputted an FT with 7 $\mathrm{BE}$s (i.e., $\pm$ $\mathrm{BE}$s $=0$), in other words, the superfluous variables were removed in the optimization process. On the other hand, the m.o.f. $d$ shows inconsistency for different values of $\rho$, and it seems to perform poorly for larger values of $\rho$.


\subsubsection{Varying the parent fault trees}\label{sec:initial_fts_strategies}

We defined in Section \ref{sec:initial_parent_fts} what a parent FT is, and here we evaluate the effects of varying this parameter. We use the case study MPPS, the m.o.f. $sdc$ with the following parameters $ps=400$, $ng=100$, $uc=20$, $\rho = 6$. We consider three setups. \textit{Setup A} is our reference, we have been using it throughout this paper (see Section \ref{sec:initial_parent_fts}). \textit{Setup B} consists of a single parent FT based on the DNF. \textit{Setup C} takes an inferred sub-optimal FT model with $\FTSize = 98$ previously obtained using the m.o.f. $d$ in Fig. \ref{fig:DU6p6C2ci0_spurious_variables}.(a).

\begin{figure}[!h]
\centering
\begin{subfigure}{.22\textwidth}
  \centering
  \includegraphics[width=1\linewidth]{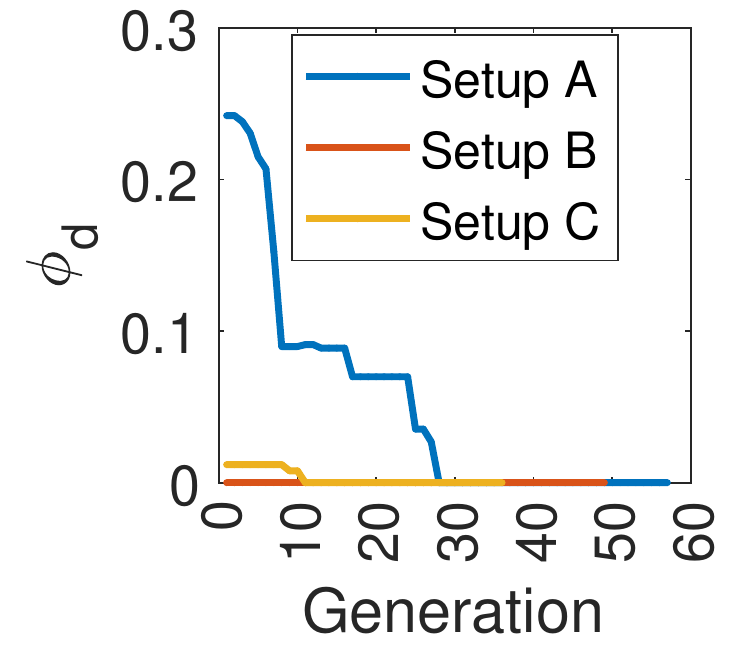}
  \caption{}
\end{subfigure}
\begin{subfigure}{.22\textwidth}
  \centering
  \includegraphics[width=1\linewidth]{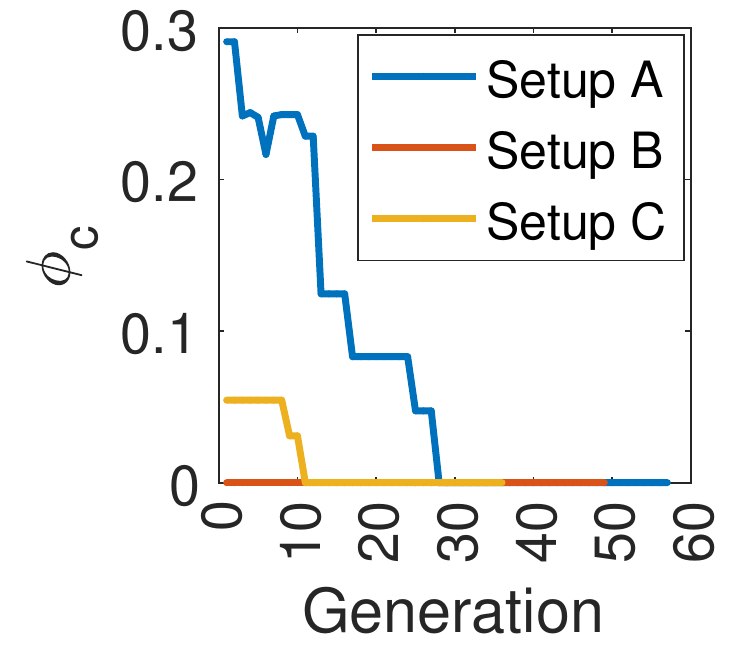}
  \caption{}
\end{subfigure}

\begin{subfigure}{.22\textwidth}
  \centering
  \includegraphics[width=1\linewidth]{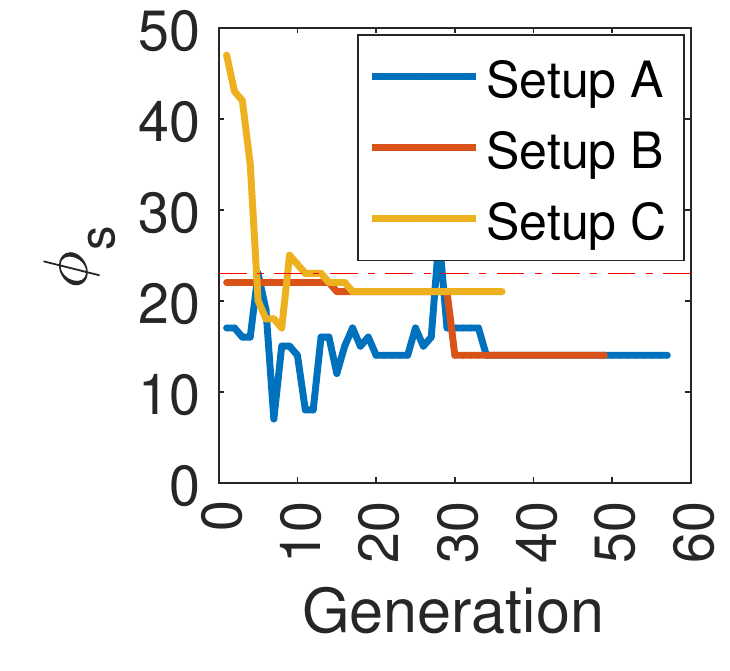}
  \caption{}
\end{subfigure}
\begin{subfigure}{.22\textwidth}
  \centering
  \includegraphics[width=1\linewidth]{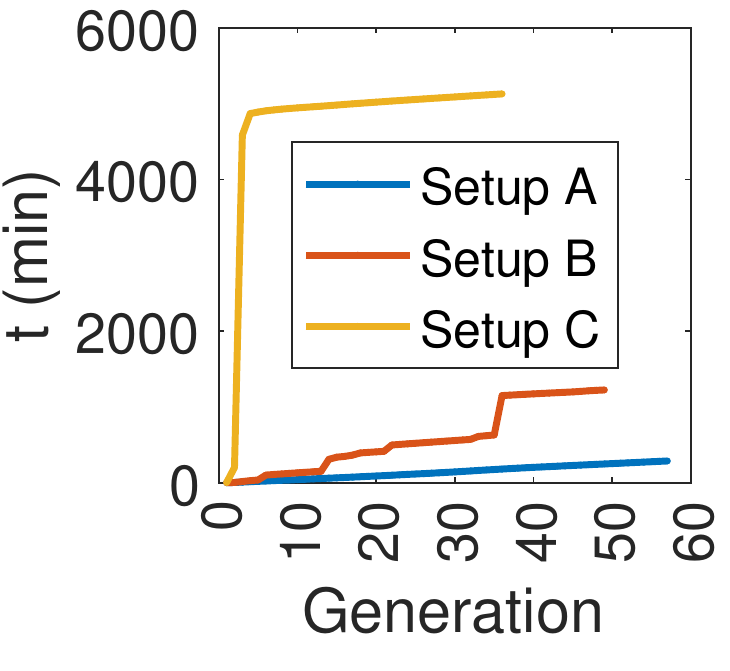}
  \caption{}
\end{subfigure}
\caption{Influence of varying the parent FTs on (a) $\ErrorData$, (b) $\ErrorData$, (c) $\FTSize$, (d) convergence time. For the m.o.f $sdc$, the case study MPPS ($ps=400$, $ng=100$, $uc=20$, $\rho = 6$). \textit{Setup A}: as described in Section \ref{sec:initial_parent_fts}; \textit{Setup B}: using as parent FT the \textit{disjunctive normal form}; \textit{Setup C}: using as parent FT a sub-optimal FT of $\phi_s = 98$ obtained with the m.o.f. $d$ in Fig. \ref{fig:DU6p6C2ci0_spurious_variables}.(d).}
\label{fig:G7AmvD7Enc_varying_initial_parent_FTs}
\end{figure}

We already discussed in detail the results of Setup A in Fig. \ref{fig:MmDSvSeoKf_output_ft_comparison_over_generations}, Section \ref{sec:multi_inside}. Fig. \ref{fig:G7AmvD7Enc_varying_initial_parent_FTs}.(a) and (b) shows that Setup B has an error of zero since the onset (i.e., $\ErrorData = \ErrorData = 0.0$). This is expected because Setup B considers as parent FT the DNF. However, this does not mean the parent FT has the optimal structure, this is what we observe in Fig. \ref{fig:G7AmvD7Enc_varying_initial_parent_FTs}.(c) where around the 30th generation the FT-MOEA found a smaller structure with the same performance. The latter structure is the same as the one where Setup A converged.

Setup C started with a sub-optimal parent FT of $\FTSize = 98$ but in Fig. \ref{fig:G7AmvD7Enc_varying_initial_parent_FTs}.(c) we observe that right from the first generation, the FT-MOEA algorithm found a better solution of $\FTSize = 47$, and it converged in $\FTSize = 21$. Fig. \ref{fig:G7AmvD7Enc_varying_initial_parent_FTs}.(d) shows that it took about 3 days to get through the first three generations, the reason being that computing the MCSs of many large FTs was expensive, but as soon as the FTs of the population reached a smaller size, the process became faster.

From the above results, we can conclude that Setup A is faster than Setups B and C, although it took more generations to find the global optimum. On the contrary, Setup C found the global optimum with fewer generations, but both, Setup B and C are slower, the reason most likely is because obtaining MCSs from large FTs is computationally expensive. 

\section{Discussion and Conclusions}\label{sec:disucssion_conclusions}

In this paper we demonstrate that it is possible to infer efficient and interpretable FT structures by implementing multi-objective evolutionary algorithms. However, there are several aspects that still need to be addressed before thinking about real-world applications.

Even though our algorithm fares better than alternative approaches, its \textit{scalability} will become an issue for FTs with many BEs. A way to overcome this drawback is by means of "\textit{guided}" \textit{multi-objective evolutionary algorithms}. This concept consists of helping the optimization process by injecting additional knowledge. One way to address this is by searching out for \textit{patterns} (e.g., Waghen \& Ouali \cite{waghen2021multi}) in the failure data set that enables identifying parts of the FT and then assemble them. 

Another way is by \textit{guiding} the application of the \textit{genetic operators}. In our current procedure these are randomly applied, whereas one may be able to increase the chances of finding the global optimum by targeting at parts of the FT that most likely need to be modified. The latter might be possible to achieve by implementing Bayesian optimization. A more challenging path is exploring deep learning-based approaches like the one by Cranmer et al. \cite{cranmer2020discovering} to derive symbolic rules from a Graph Neural Networks. 

We found that the inclusion of MCSs in the multi-objective optimization function greatly improves the optimization process. However, the FT-MOEA computes the MCSs based on the disjunctive normal form, which is computationally expensive for large FTs. Moreover, MCSs cannot be computed in noisy data. Thus, alternatives to overcome these issues are worth exploring. Finally, some spare but equally interesting challenges are enlisted next.

\begin{itemize}

    \item Having complete failure data sets for complex engineering systems is virtually impossible. Thus, a more thorough evaluation of the performance of our algorithm under incomplete failure data sets is necessary.
    
    \item Real-world problems often contain symmetries, e.g., when two basic events are known to be fully exchangeable. This property could be used as an advantage, because we expect to reduce the solution space, leading to faster convergence. Thus, research in this direction is needed.
    
    \item In order to obtain more compact and efficient FT structures, exploring alternatives that enable the inference of more sophisticated gates (e.g., VoT gates) is necessary.
    
    
    \item We believe that a methodology similar to the one used in this paper could be applied for the inference of other reliability models, such as reliability block diagrams, as well as Boolean circuits.
    
\end{itemize}

Our novel algorithm, FT-MOEA, has in general a better performance than its predecessor, the FT-GA, by converging faster, inferring more compact FT structures, achieving lower error levels, better removing superfluous variables, and being consistent.
\ifCLASSOPTIONcompsoc
  \section*{Acknowledgments}
\else
  \section*{Acknowledgment}
\fi

Thanks to Matthias Volk for providing us with useful advice. This research has been partially funded by NWO under the grant PrimaVera (https://primavera-project.com) number NWA.1160.18.238.

\ifCLASSOPTIONcaptionsoff
  \newpage
\fi




%

\bibliographystyle{IEEEtran}
\bibliography{references}

\begin{thebibliography}{10}
\providecommand{\url}[1]{#1}
\csname url@samestyle\endcsname
\providecommand{\newblock}{\relax}
\providecommand{\bibinfo}[2]{#2}
\providecommand{\BIBentrySTDinterwordspacing}{\spaceskip=0pt\relax}
\providecommand{\BIBentryALTinterwordstretchfactor}{4}
\providecommand{\BIBentryALTinterwordspacing}{\spaceskip=\fontdimen2\font plus
\BIBentryALTinterwordstretchfactor\fontdimen3\font minus
  \fontdimen4\font\relax}
\providecommand{\BIBforeignlanguage}[2]{{%
\expandafter\ifx\csname l@#1\endcsname\relax
\typeout{** WARNING: IEEEtran.bst: No hyphenation pattern has been}%
\typeout{** loaded for the language `#1'. Using the pattern for}%
\typeout{** the default language instead.}%
\else
\language=\csname l@#1\endcsname
\fi
#2}}
\providecommand{\BIBdecl}{\relax}
\BIBdecl

\bibitem{kabir2017overview}
S.~Kabir, ``An overview of fault tree analysis and its application in model
  based dependability analysis,'' \emph{Expert Systems with Applications},
  vol.~77, pp. 114--135, 2017.

\bibitem{signoret2021automated}
J.-P. Signoret, A.~Leroy \emph{et~al.}, ``Automated fault tree building,''
  \emph{Springer Series in Reliability Engineering}, pp. 423--426, 2021.

\bibitem{salem1976computer}
S.~L. Salem, G.~Apostolakis, and D.~Okrent, ``Computer-oriented approach to
  fault-tree construction,'' California Univ., Tech. Rep., 1976.

\bibitem{hunt1993propagation}
A.~Hunt, B.~Kelly, J.~Mullhi, F.~Lees, and A.~Rushton, ``The propagation of
  faults in process plants: 6, overview of, and modelling for, fault tree
  synthesis,'' \emph{Reliability Engineering \& System Safety}, vol.~39, no.~2,
  pp. 173--194, 1993.

\bibitem{madden1970generation}
M.~G. Madden and P.~J. Nolan, ``Generation of fault trees from simulated
  incipient fault case data,'' \emph{WIT Transactions on Information and
  Communication Technologies}, vol.~6, 1994.

\bibitem{latif2002comparing}
G.~Latif-Shabgahi, ``Comparing selected knowledge-based fault tree construction
  tools,'' in \emph{Proc. IASTED Int. Conf. Intell. Syst. Control}, 2002.

\bibitem{carpignano1994computer}
A.~Carpignano and A.~Poucet, ``Computer assisted fault tree construction: a
  review of methods and concerns,'' \emph{Reliability Engineering \& System
  Safety}, vol.~44, no.~3, pp. 265--278, 1994.

\bibitem{mhenni2014automatic}
F.~Mhenni, N.~Nguyen, and J.-Y. Choley, ``Automatic fault tree generation from
  sysml system models,'' in \emph{2014 IEEE/ASME International Conference on
  Advanced Intelligent Mechatronics}.\hskip 1em plus 0.5em minus 0.4em\relax
  IEEE, 2014, pp. 715--720.

\bibitem{dickerson2018formal}
C.~E. Dickerson, R.~Roslan, and S.~Ji, ``A formal transformation method for
  automated fault tree generation from a uml activity model,'' \emph{IEEE
  Transactions on Reliability}, vol.~67, no.~3, pp. 1219--1236, 2018.

\bibitem{quinlan1986induction}
J.~R. Quinlan, ``Induction of decision trees,'' \emph{Machine learning},
  vol.~1, no.~1, pp. 81--106, 1986.

\bibitem{madden1970hierarchically}
M.~G. Madden, ``Hierarchically structured inductive learning for fault
  diagnosis,'' \emph{WIT Transactions on Information and Communication
  Technologies}, vol.~20, 1998.

\bibitem{madden1999monitoring}
M.~G. Madden and P.~J. Nolan, ``Monitoring and diagnosis of multiple incipient
  faults using fault tree induction,'' \emph{IEE Proceedings-Control Theory and
  Applications}, vol. 146, no.~2, pp. 204--212, 1999.

\bibitem{mukherjee2007automated}
S.~Mukherjee and A.~Chakraborty, ``Automated fault tree generation: bridging
  reliability with text mining,'' in \emph{2007 Annual Reliability and
  Maintainability Symposium}.\hskip 1em plus 0.5em minus 0.4em\relax IEEE,
  2007, pp. 83--88.

\bibitem{roth2015integrated}
M.~Roth, M.~Wolf, and U.~Lindemann, ``Integrated matrix-based fault tree
  generation and evaluation,'' \emph{Procedia Computer Science}, vol.~44, pp.
  599--608, 2015.

\bibitem{li2016causal}
J.~Li, S.~Ma, T.~Le, L.~Liu, and J.~Liu, ``Causal decision trees,'' \emph{IEEE
  Transactions on Knowledge and Data Engineering}, vol.~29, no.~2, pp.
  257--271, 2016.

\bibitem{nauta2018lift}
M.~Nauta, D.~Bucur, and M.~Stoelinga, ``Lift: Learning fault trees from
  observational data,'' in \emph{International Conference on Quantitative
  Evaluation of Systems}.\hskip 1em plus 0.5em minus 0.4em\relax Springer,
  2018, pp. 306--322.

\bibitem{waghen2019interpretable}
K.~Waghen and M.-S. Ouali, ``Interpretable logic tree analysis: A data-driven
  fault tree methodology for causality analysis,'' \emph{Expert Systems with
  Applications}, vol. 136, pp. 376--391, 2019.

\bibitem{linard2019induction}
A.~Linard, M.~L. Bueno, D.~Bucur, and M.~Stoelinga, ``Induction of fault trees
  through bayesian networks,'' 09 2019.

\bibitem{Feng2020DATADRIVENFT}
S.~Lazarova-Molnar, P.~Niloofar, and G.~K. Barta, ``Data-driven fault tree
  modeling for reliability assessment of cyber-physical systems,'' in
  \emph{Proceedings of the 2020 Winter Simulation Conference}, 2020.

\bibitem{waghen2021multi}
K.~Waghen and M.-S. Ouali, ``Multi-level interpretable logic tree analysis: A
  data-driven approach for hierarchical causality analysis,'' \emph{Expert
  Systems with Applications}, vol. 178, p. 115035, 2021.

\bibitem{linard2019fault}
A.~Linard, D.~Bucur, and M.~Stoelinga, ``Fault trees from data: Efficient
  learning with an evolutionary algorithm,'' in \emph{International Symposium
  on Dependable Software Engineering: Theories, Tools, and Applications}.\hskip
  1em plus 0.5em minus 0.4em\relax Springer, 2019, pp. 19--37.

\bibitem{deb2014multi}
K.~Deb, ``Multi-objective optimization,'' in \emph{Search methodologies}.\hskip
  1em plus 0.5em minus 0.4em\relax Springer, 2014, pp. 403--449.

\bibitem{ruijters2015fault}
E.~Ruijters and M.~Stoelinga, ``Fault tree analysis: A survey of the
  state-of-the-art in modeling, analysis and tools,'' \emph{Computer science
  review}, vol.~15, pp. 29--62, 2015.

\bibitem{stamatelatos2002fault}
M.~Stamatelatos, W.~Vesely, J.~Dugan, J.~Fragola, J.~Minarick, and
  J.~Railsback, ``Fault tree handbook with aerospace applications,'' 2002.

\bibitem{ojha2019review}
M.~Ojha, K.~P. Singh, P.~Chakraborty, and S.~Verma, ``A review of
  multi-objective optimisation and decision making using evolutionary
  algorithms,'' \emph{International Journal of Bio-Inspired Computation},
  vol.~14, no.~2, pp. 69--84, 2019.

\bibitem{deb2011multi}
K.~Deb, ``Multi-objective optimisation using evolutionary algorithms: an
  introduction,'' in \emph{Multi-objective evolutionary optimisation for
  product design and manufacturing}.\hskip 1em plus 0.5em minus 0.4em\relax
  Springer, 2011, pp. 3--34.

\bibitem{deb2002fast}
K.~Deb, A.~Pratap, S.~Agarwal, and T.~Meyarivan, ``A fast and elitist
  multiobjective genetic algorithm: Nsga-ii,'' \emph{IEEE transactions on
  evolutionary computation}, vol.~6, no.~2, pp. 182--197, 2002.

\bibitem{long2021non}
Q.~Long, X.~Wu, and C.~Wu, ``Non-dominated sorting methods for multi-objective
  optimization: Review and numerical comparison,'' \emph{Journal of Industrial
  \& Management Optimization}, vol.~17, no.~2, p. 1001, 2021.

\bibitem{marti2017impact}
L.~Marti, E.~Segredo, E.~Hart \emph{et~al.}, ``Impact of selection methods on
  the diversity of many-objective pareto set approximations,'' \emph{Procedia
  computer science}, vol. 112, pp. 844--853, 2017.

\bibitem{robert1976unifying}
P.~Robert and Y.~Escoufier, ``A unifying tool for linear multivariate
  statistical methods: The rv-coefficient,'' \emph{Journal of the Royal
  Statistical Society: Series C (Applied Statistics)}, vol.~25, no.~3, pp.
  257--265, 1976.

\bibitem{bakeli2020covid}
T.~Bakeli, A.~A. Hafidi \emph{et~al.}, ``Covid-19 infection risk management
  during construction activities: An approach based on fault tree analysis
  (fta),'' \emph{Journal of Emergency Management}, vol.~18, no.~7, pp.
  161--176, 2020.

\bibitem{mentes2011application}
A.~Mentes and I.~H. Helvacioglu, ``An application of fuzzy fault tree analysis
  for spread mooring systems,'' \emph{Ocean Engineering}, vol.~38, no. 2-3, pp.
  285--294, 2011.

\bibitem{cranmer2020discovering}
M.~Cranmer, A.~Sanchez-Gonzalez, P.~Battaglia, R.~Xu, K.~Cranmer, D.~Spergel,
  and S.~Ho, ``Discovering symbolic models from deep learning with inductive
  biases,'' \emph{arXiv preprint arXiv:2006.11287}, 2020.

\bibitem{berikov2004fault}
V.~Berikov, ``Fault tree construction on the basis of multivariate time series
  analysis,'' in \emph{Proceedings. The 8th Russian-Korean International
  Symposium on Science and Technology, 2004. KORUS 2004.}, vol.~2.\hskip 1em
  plus 0.5em minus 0.4em\relax IEEE, 2004, pp. 103--106.

\end{thebibliography}




%

\vspace{-10 mm}

\begin{IEEEbiography}[{\includegraphics[width=1in,height=1.25in,clip,keepaspectratio]{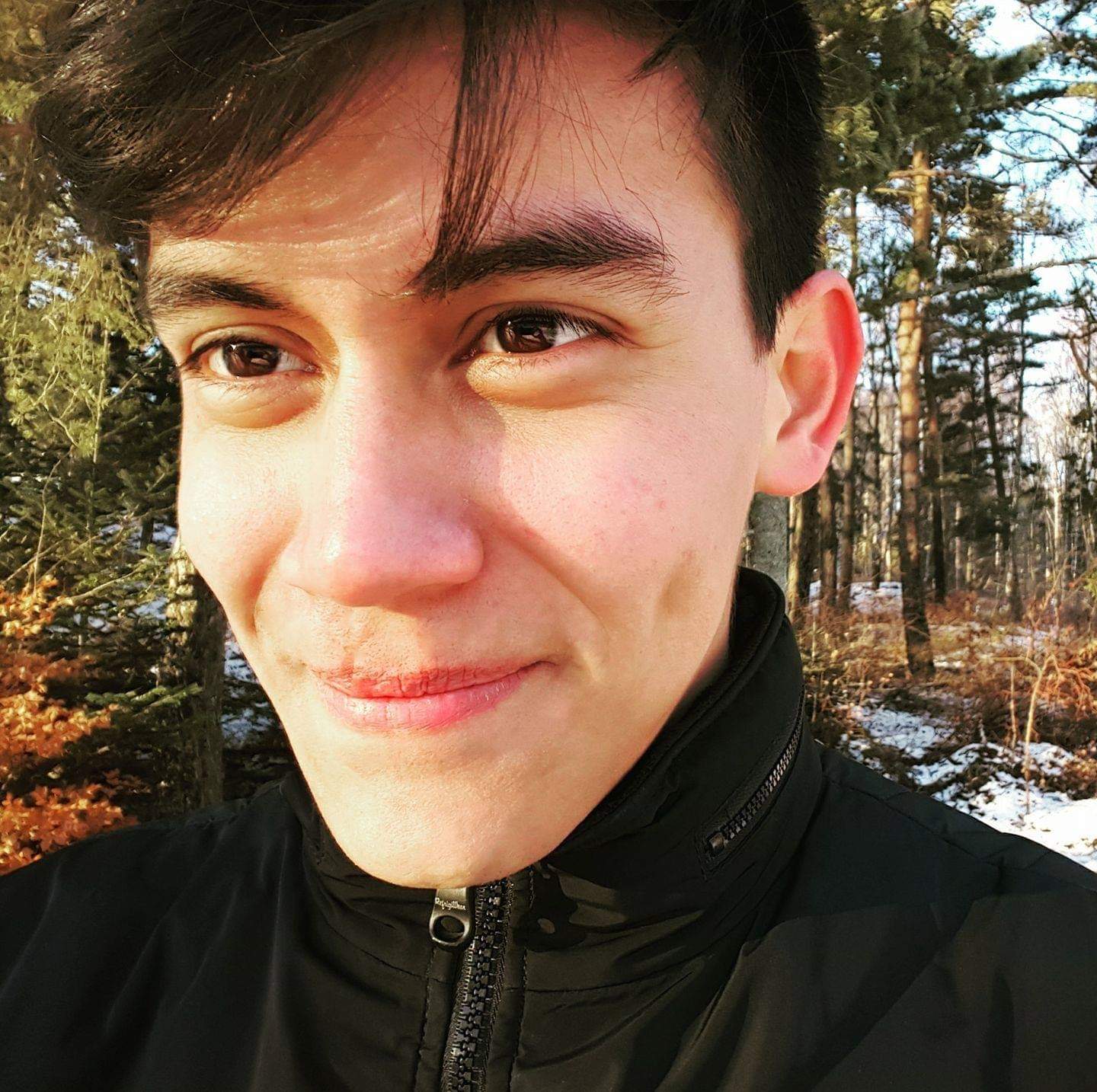}}]{Lisandro A. Jimenez-Roa} is a doctorate candidate in computer science at the University of Twente, The Netherlands. His background is in civil engineering and have worked in a variety of projects on the fields of structural health monitoring, finite element modelling, and damage detection via data analytics and machine learning. Currently, he is conducting research on system-level prognostics of complex engineering systems within the PrimaVera project (https://primavera-project.com), with particular interest on hybrid integration of domain expertise, physics-informed and data-driven approaches.
\end{IEEEbiography}

\vspace{-10 mm}

\begin{IEEEbiography}[{\includegraphics[width=1in,height=1.25in,clip,keepaspectratio]{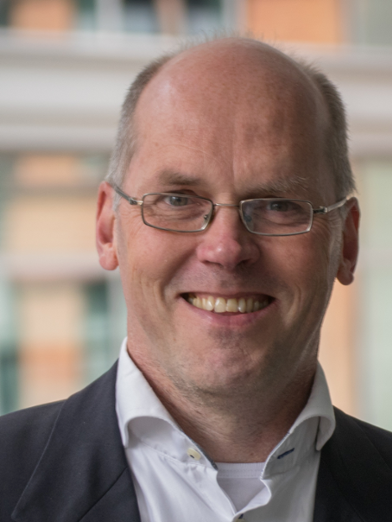}}]{Tom Heskes} is full professor of Artificial Intelligence. After receiving his PhD on neural networks, he worked as a postdoc at the Beckman Institute in Champaign-Urbana, Illinois. Back in the Netherlands, he joined SNN, the Foundation for Neural Networks and later the Institute for Computing and Information Sciences at Radboud University. Tom Heskes is former Editor-in-Chief of Neurocomputing and has been (co-)leading various national and European projects. Heskes’ research concerns the development, understanding, and application of machine learning methods, currently in particular deep learning and causal inference. He works on applications in other scientific disciplines, as well as in industry, among others through his spin-off company Machine2Learn.
\end{IEEEbiography}

\vspace{-10 mm}

\begin{IEEEbiography}[{\includegraphics[width=1in,height=1.25in,clip,keepaspectratio]{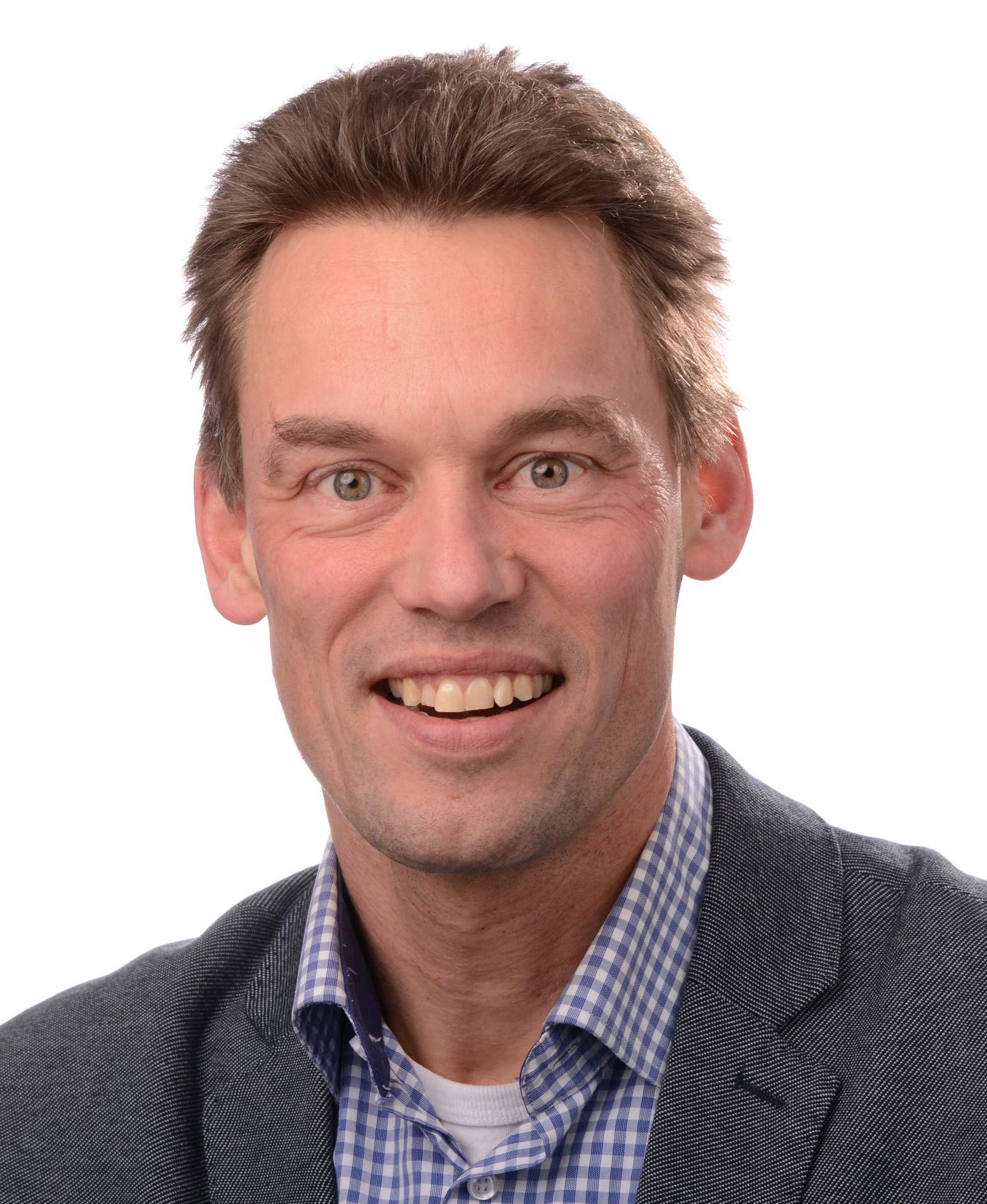}}]{Tiedo Tinga}
is a full professor in dynamics based maintenance at the University of Twente since 2012 and full professor Life Cycle Management at the Netherlands Defence
Academy since 2016. He received his PhD degree in mechanics of materials from Eindhoven University in 2009. He is chairing the smart maintenance knowledge center and
leads a number of research projects on developing predictive maintenance concepts, mainly based on physics of failure models, but also following data-driven approaches.
\end{IEEEbiography}

\vspace{-10 mm}

\begin{IEEEbiography}[{\includegraphics[width=1in,height=1.25in,clip,keepaspectratio]{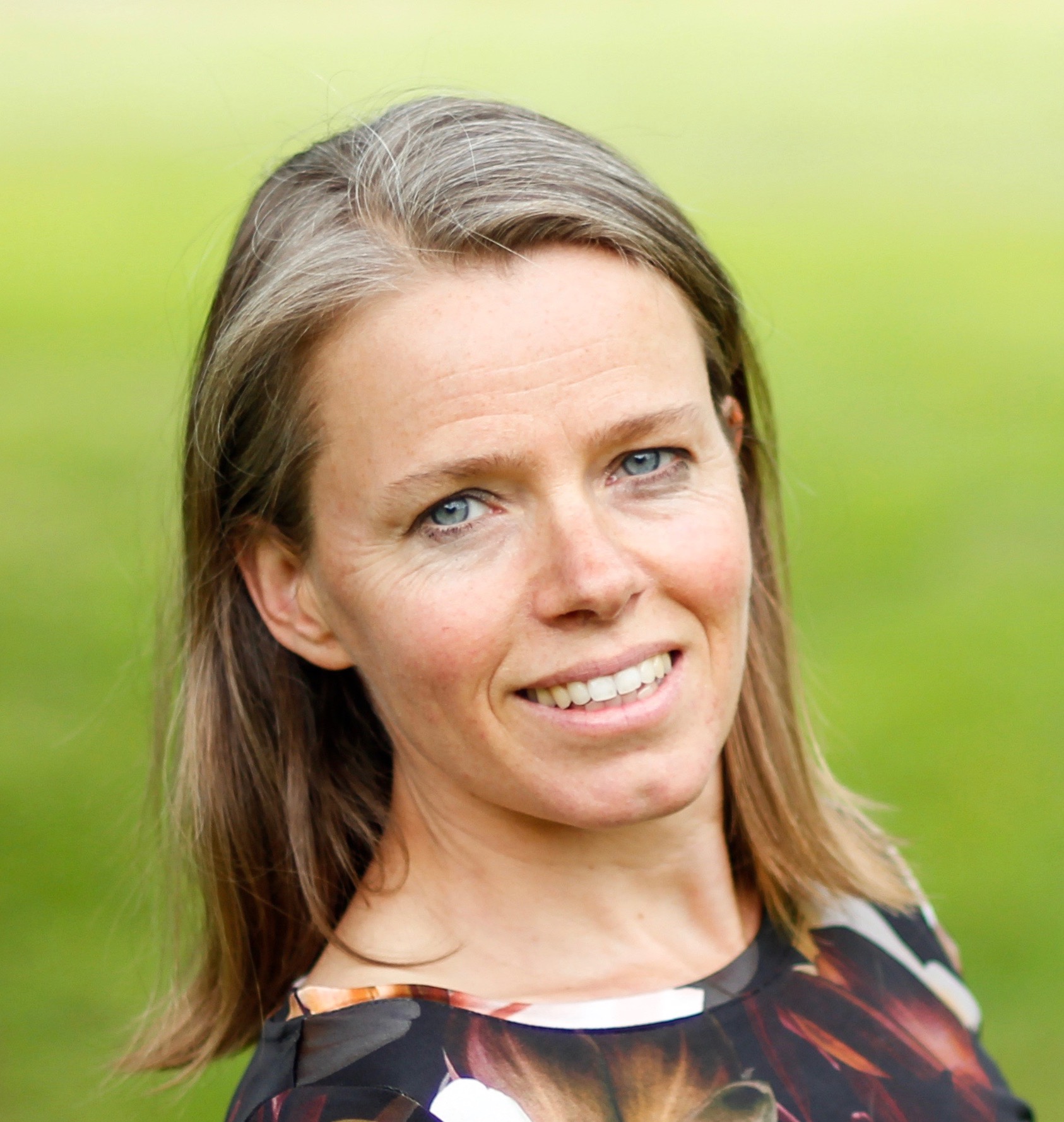}}]{Marielle Stoelinga}
is a full professor of risk analysis for high-tech systems, both at the University of Twente and Radboud University, the Netherlands. She holds a Master's degree in Mathematics \& Computer Science, and a PhD in Computer Science. After her PhD, she has been a postdoctoral researcher at the University of California at Santa Cruz, USA. 

Prof. Stoelinga leads the executive Master on Risk Management at the University of Twente, a part time programme for risk professionals. She also leads various research projects, including a large national consortium on Predictive Maintenance and an ERC consolidator grant on safety and security interactions.

\end{IEEEbiography}


\vfill

\newpage


\appendices

\section{Applying NSGA-II and Crowding-Distance to infer FTs}\label{app:application_MOEAs_to_FTs}

\subsection{Applying NSGA-II to infer FTs}\label{sec:app_nsgaII_to_infer_fts}

\begin{figure}[!h]
\centering
\includegraphics[width=0.9\linewidth]{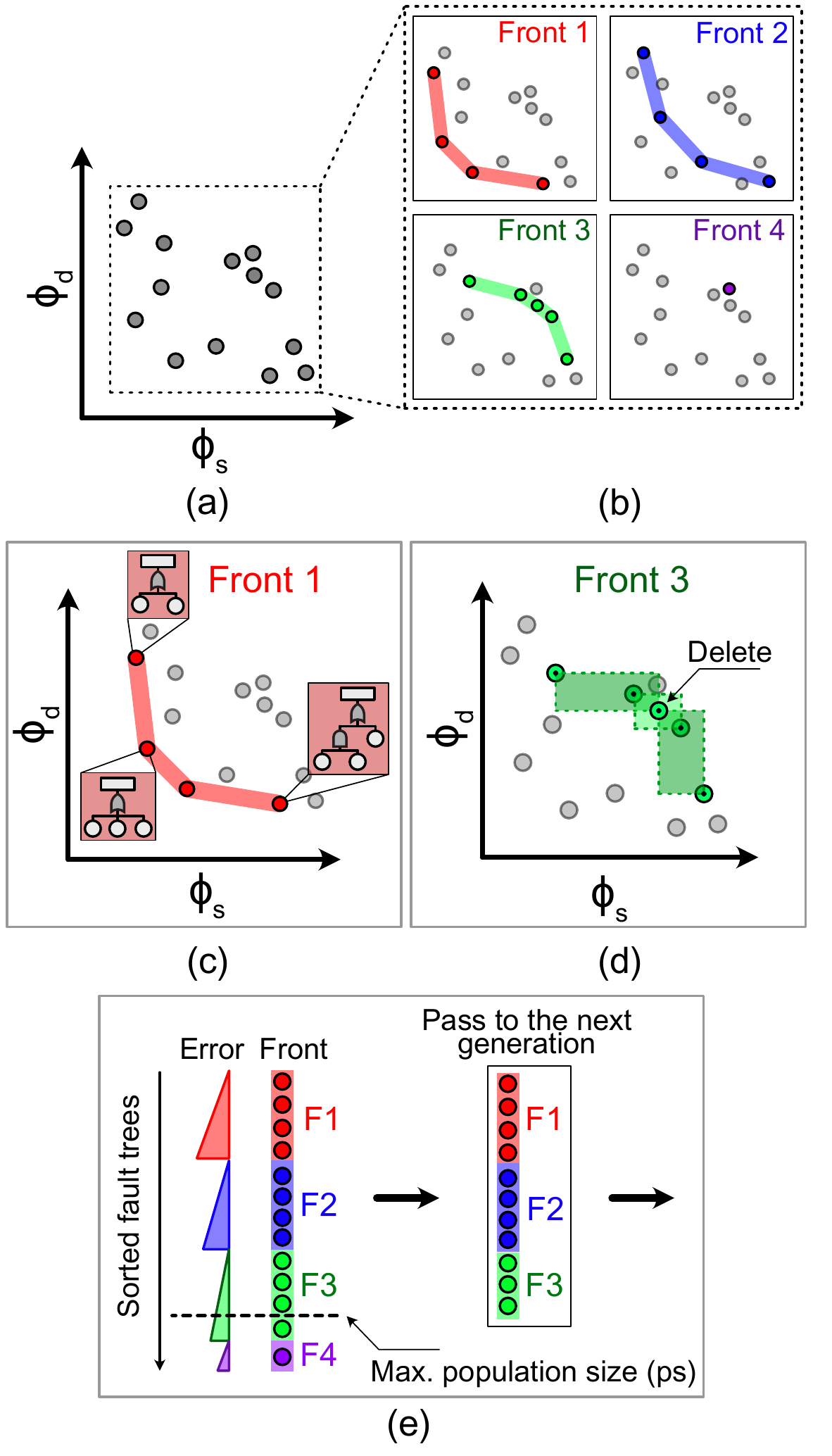}
\caption{Conceptual visualization of the \textit{Non-Dominated Sorting Genetic Algorithms} (NSGA-II) and \textit{Crowding-Distance} in the context of automatic inference of FTs. In (a) error based on the failure data set ($\ErrorData$) versus fault tree size ($\FTSize$), in (b) Pareto fronts, (c) details of the first front, (d) influence of the Crowding-Distance, and (e) criteria for acceptance and rejection of FTs between generations.}
\label{fig:cOV8l0MbZj_fig_NSGA_explanation}
\end{figure}

We provide a conceptual visualisation in Fig. \ref{fig:cOV8l0MbZj_fig_NSGA_explanation} that explains our implementation of the NSGA-II and Crowding-Distance in the context of the automatic inference of FTs. To ease the visualisation, we consider the bi-dimensional case where the multi-objective function is \textit{sd} (see Table \ref{tb:setups_mof}). After computing the metrics for a population of FTs within a given generation, one can depict the FTs with circles as in Fig. \ref{fig:cOV8l0MbZj_fig_NSGA_explanation}.(a). 

The output of the NSGA-II algorithm is a set of \textit{Pareto fronts}, represented in Fig. \ref{fig:cOV8l0MbZj_fig_NSGA_explanation}.(b) with different colours (red, blue, green and purple). Fig. \ref{fig:cOV8l0MbZj_fig_NSGA_explanation}.(c) shows some details related to the FTs in the first front (red). Note that these FTs have different structures. Here the top FT has a higher error based on the failure data set ($\ErrorData$) compared to the others, but it is the smallest FT in the first front. On the contrary, the bottom FT has a smaller error in $\ErrorData$, but with the trade-off of having a larger size. 

Fig. \ref{fig:cOV8l0MbZj_fig_NSGA_explanation}.(d) shows the effect of the \textit{Crowding-Distance}. Suppose that only four of the five FTs of the third front can pass to the next generation. Therefore, it is necessary to ``break'' the front. To do so, we compute the \textit{Crowding-Distance metric} ($d_i$) (Section \ref{sec:app_crowding_distance}). Those solutions in the front with a large $d_i$ value have priority to pass to the next generation. Conversely, those with a small $d_i$ value have a lower priority because it means that they are similar to other solutions in the front. Therefore, in Fig. \ref{fig:cOV8l0MbZj_fig_NSGA_explanation}.(d) the FT marked with the arrow ``Delete'' must have similar features in $\FTSize$ and $\ErrorData$ compared to its neighbours, thus becoming the candidate to be deleted from the third front.

Figure \ref{fig:cOV8l0MbZj_fig_NSGA_explanation}.(e) represents the process to select the FTs that pass to the next generation, where first the FTs within each front are ordered from the minimum to the maximum error (or errors when considering the minimisation of both $\phi_{c}$ and $\phi_{d}$, here we sum them up and sort them up), then only the first $ps$ FTs pass to the next generation. Here we can observe that one FT from the third front and the only FT from the fourth front did not manage to pass to the next generation.

\subsection{Crowding-Distance}\label{sec:app_crowding_distance}
This process is based on the \textit{Crowding-Distance metric} ($d_i$) which makes that individuals with a large $d_i$ wins, the latter to avoid solutions to be similar (i.e., to maintain diversity). This metric estimates the density of a particular solution in the non-dominated front based on the neighbour solutions. We provide below a summary of the steps necessary to compute $d_i$. For details, we suggest the reader to consult Deb et al. \cite{deb2014multi}.

\begin{itemize}
	\item Step 1: For each solution in the non-dominated set ($\mathscr{F}$) assign $d_i = 0$.
	
	\item Step 2: for all the objective functions $m=1,2,...,M$, sort the set in worse order of $f_m$ and obtain the sorted index vector as $I^m = sort(f_m,>)$.
	
	\item Step 3: For all the objective functions, assign a large distance to the boundary solutions (i.e., $d_{I_1^m} = d_{I_{|F|}^m} = \infty $), and for the rest of the solutions $j=2$ to $(|\mathscr{F}| - 1)$ , assign
	\begin{equation}
		d_{I_j^m} = d_{I_j^m} + \frac{f_m^{I^m_{j+1}}-f_m^{I^m_{j-1}}}{f_m^{max}-f_m^{min}}
	\end{equation}	

	Here $I_j$ denotes the solution index of the $j$-th member sorted in the vector.
	
	\item Step 4: Pass the solution with the largest $d_i$, then the solution with the second largest $d_i$, until the maximum population size is met.
\end{itemize}


\section{Example of inferred fault trees}\label{app:example_output_fts}

\begin{figure}[!h]
\centering
\includegraphics[width=1\linewidth]{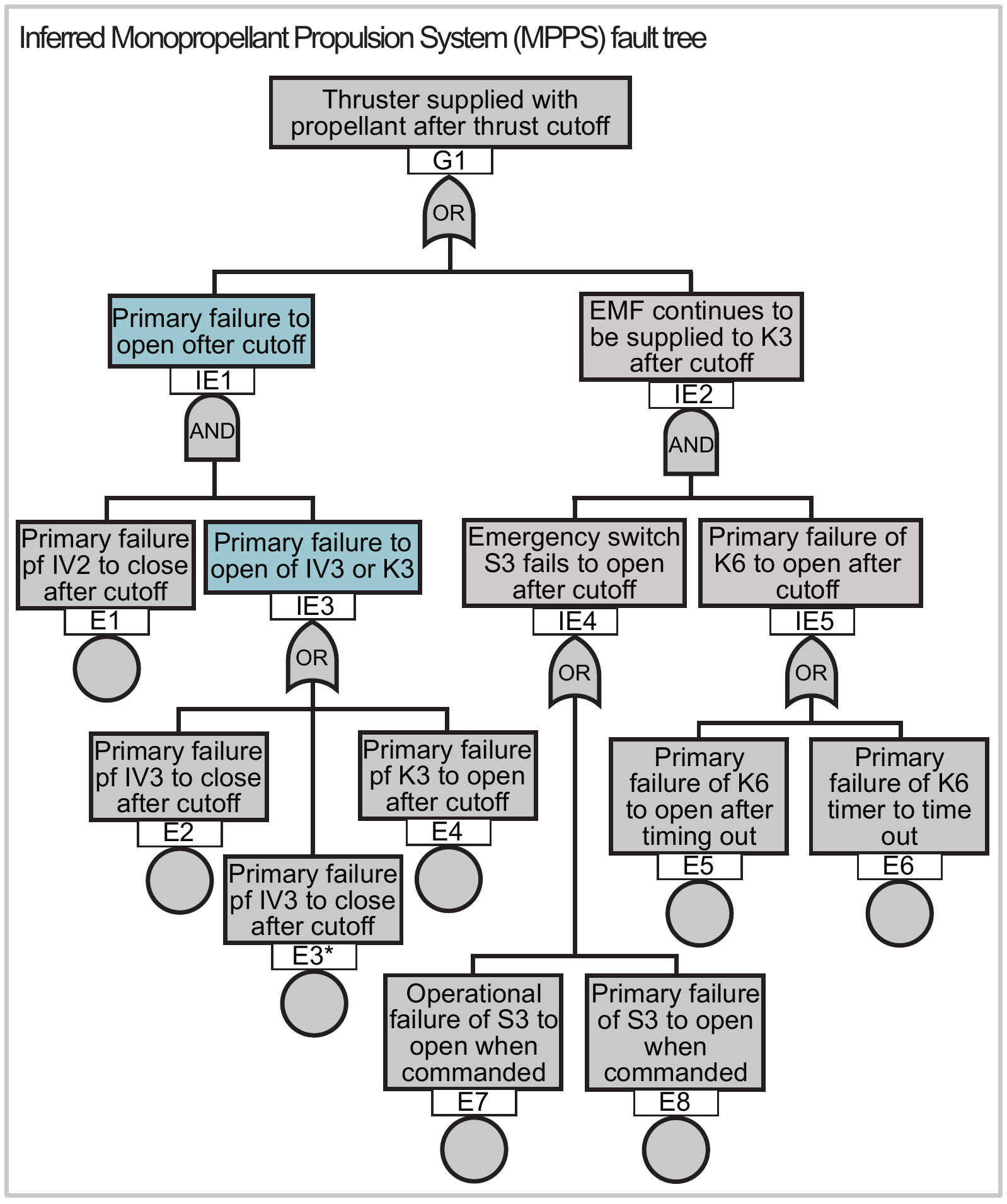}
\caption{Inferred Mono-propellant Propulsion System (MPPS) fault tree after applying FT-MOEA, source: Stamatelatos et al. \cite{stamatelatos2002fault}.}
\label{fig:BF7w7dVZI1_MPPS_fault_tree}
\end{figure}

Here we provide as an example a pair of FTs obtained with the FT-MOEA. Fig. \ref{fig:BF7w7dVZI1_MPPS_fault_tree} shows the inferred FT associated with the \textit{Monopropellant Propulsion System} (MPPS) case study, the example that was used several times to exemplify our results. Here the right branch of the FT (i.e., the one associated with the intermediate event IE2), both sets of $\mathrm{BE}$s and $\mathrm{IE}$s coincide with the ground truth FT. In contrast, the intermediate events on the left branch were inferred by the FT-MOEA.

\begin{figure}[!h]
\centering
\includegraphics[width=1\linewidth]{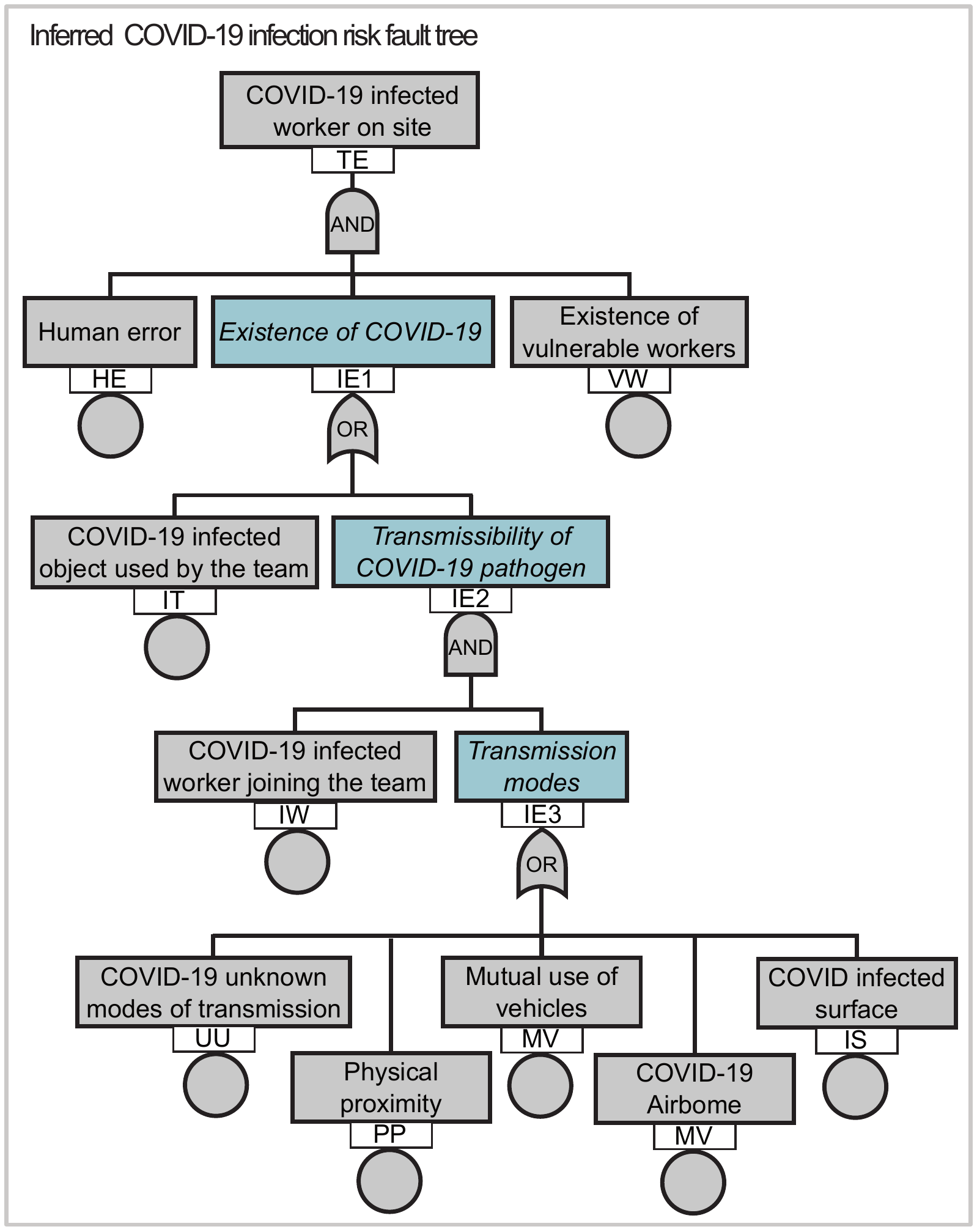}
\caption{Inferred COVID-19 infection risk fault tree after applying FT-MOEA, source: Bakeli et al. \cite{bakeli2020covid}.}
\label{fig:GxTKHCOfPh_covid_19_reduced}
\end{figure}

Similarly, in Fig. \ref{fig:GxTKHCOfPh_covid_19_reduced} we present the inferred FT of the COVID- 19 infection risk. This FT originally has 33 elements, but after applying FT-MOEA, most of the intermediate events were replaced by more efficient logics, resulting in a FT with 13 elements. Here, it is necessary to provide an interpretation to the intermediate events found by the FT-MOEA (blue boxes). First, it is interesting to see that all \textit{transmission modes} were clustered under an $\mathrm{Or}$ gate ($\mathrm{IE3}$). We interpret the other two intermediate as \textit{Transmissibility of COVID-19 pathogen} ($\mathrm{And}$ gate, $\mathrm{IE2}$) and \textit{Existence of COVID-19} ($\mathrm{Or}$ gate, $\mathrm{IE1}$).

\section{Details of convergence of metrics over generations}\label{app:convergence_over_generations}

\begin{figure}[!h]
\centering
\begin{subfigure}{.228\textwidth}
  \centering
  \includegraphics[width=1\linewidth]{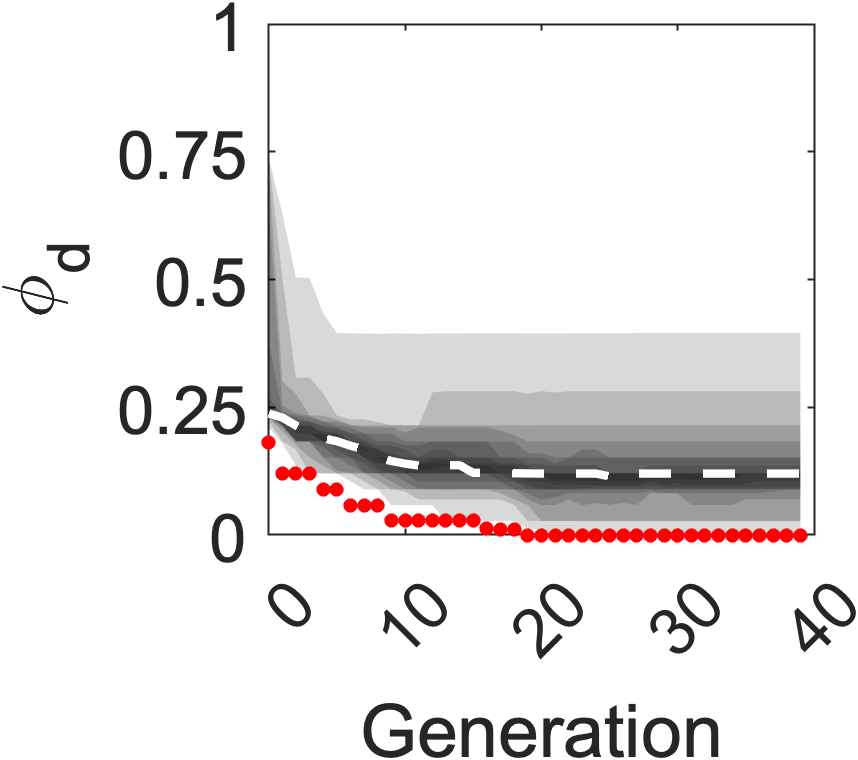}
  \caption{}
\end{subfigure}%
\begin{subfigure}{.228\textwidth}
  \centering
  \includegraphics[width=1\linewidth]{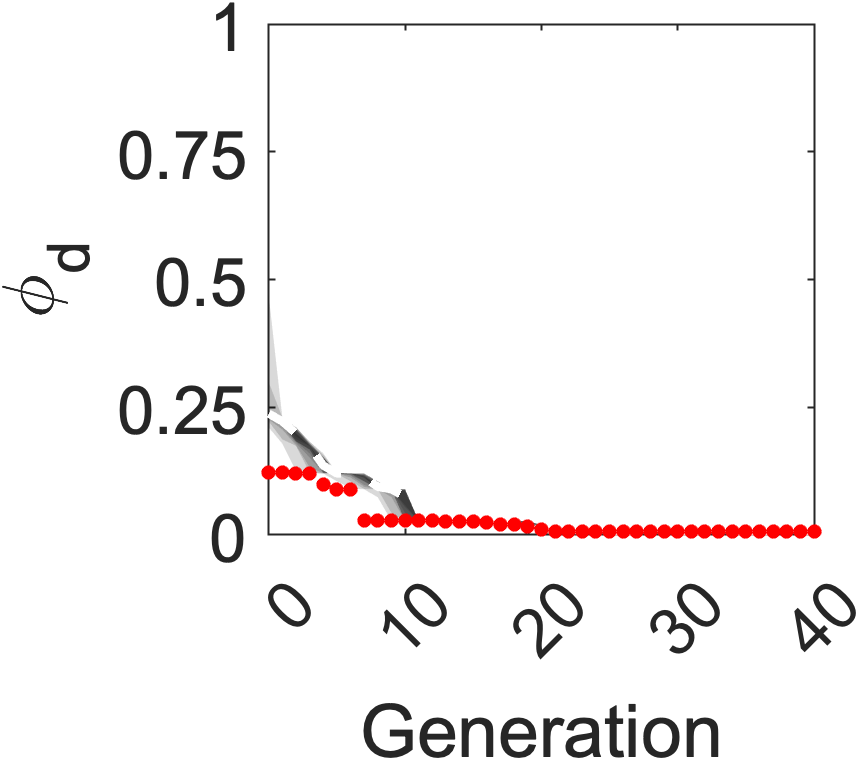}
  \caption{}
\end{subfigure}

\begin{subfigure}{.228\textwidth}
  \centering
    \includegraphics[width=1\linewidth]{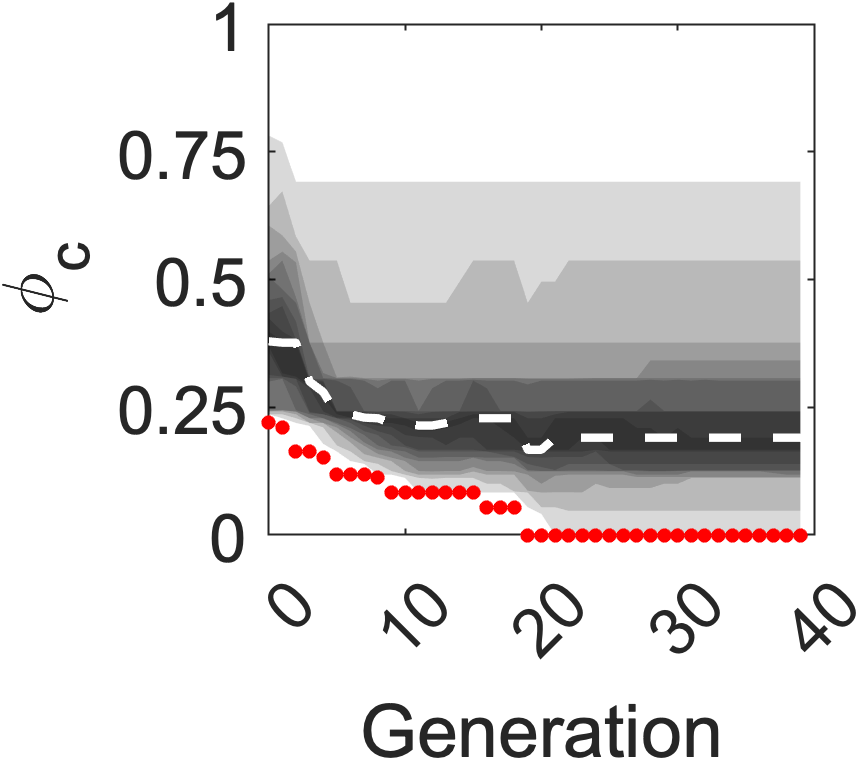}
  \caption{}
\end{subfigure}
\begin{subfigure}{.228\textwidth}
  \centering
  \includegraphics[width=1\linewidth]{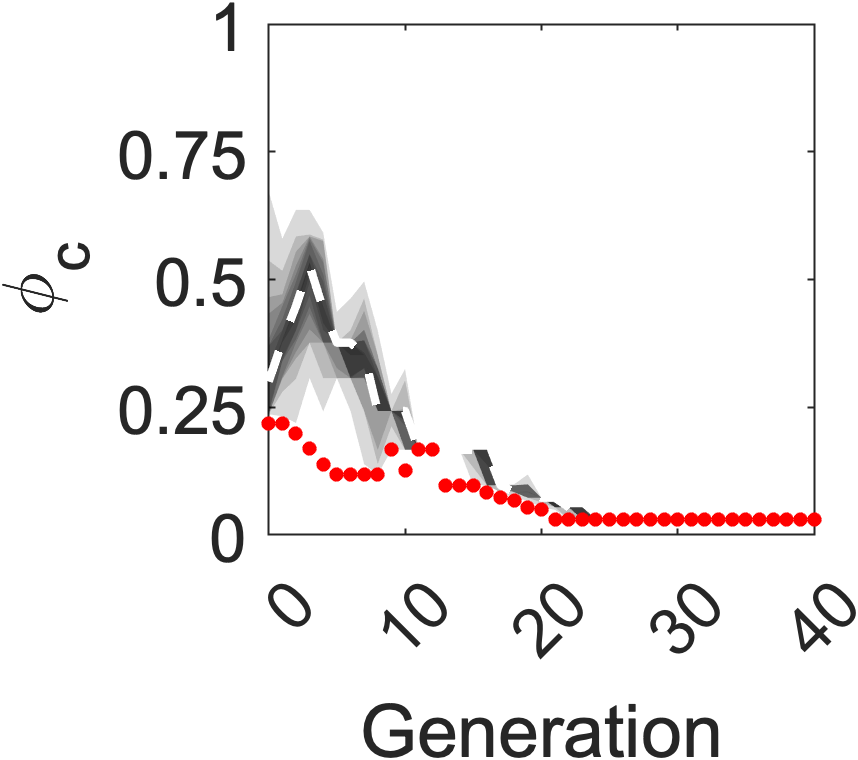}
  \caption{}
\end{subfigure}

\begin{subfigure}{.228\textwidth}
  \centering
  \includegraphics[width=1\linewidth]{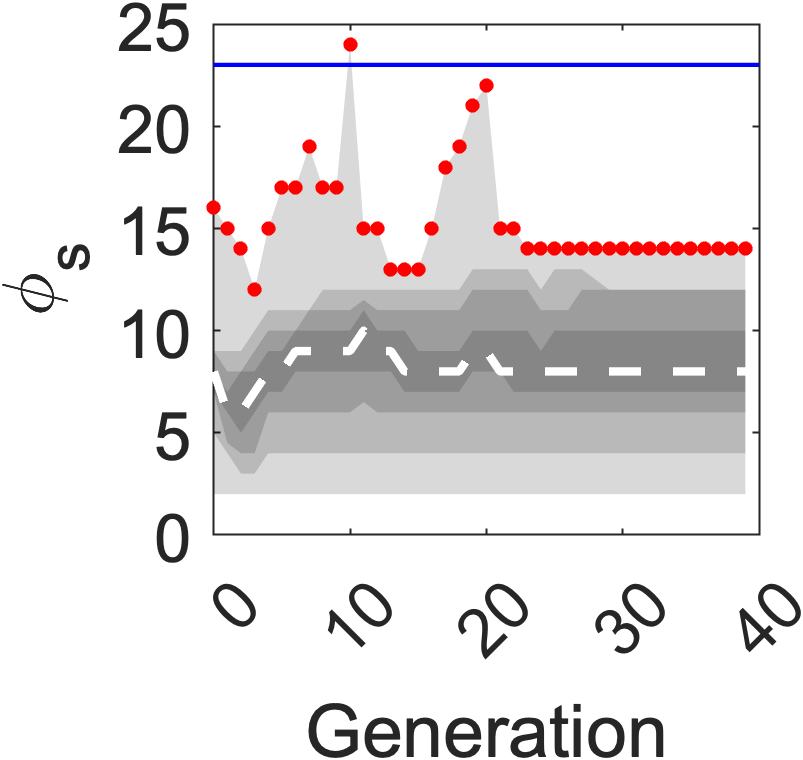}
  \caption{}
\end{subfigure}%
\begin{subfigure}{.228\textwidth}
  \centering
  \includegraphics[width=1\linewidth]{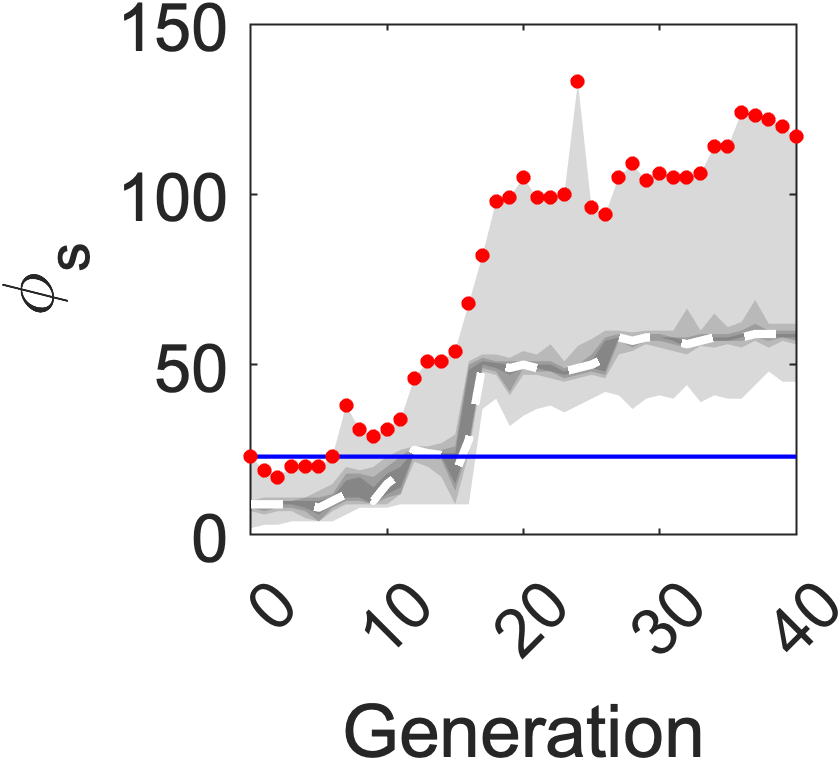}
  \caption{}
\end{subfigure}
\caption{Visualization of metrics ($\FTSize$, $\ErrorData$, $\ErrorMCSs$) over the generations considering the percentiles 25\%, 50\%, 75\% and 100\% for the case study MPPS ($ps=400$, $ng=100$, $uc=20$). In (a), (c) and (e) using the m.o.f. $sdc$, and in (b), (d), and (f) using the m.o.f. $d$.}
\label{fig:XXK0VJARWR_convergence_metrics_over_generations_all_pop}
\end{figure}

Fig. \ref{fig:XXK0VJARWR_convergence_metrics_over_generations_all_pop} depicts the convergence of the metrics $\FTSize$, $\ErrorData$, and $\ErrorMCSs$, over the generations for the same example discussed in Fig. \ref{fig:MmDSvSeoKf_output_ft_comparison_over_generations} in Section \ref{sec:multi_inside}. But now also the \textit{distribution} inside the population of each generation is visualised. The aim is to provide the reader a better understanding of what is happening between the generations with respect to the convergence of the metrics. The common elements in these figures are grey shades associated to the percentiles 25\% (darkest shade), 50\%, 75\% and 100\% (lightest shade), a horizontal blue line that indicates the size of the ground truth FT, red dots that indicate the extreme values in that generation for a given metric, and a white dashed line that indicates the mean value of the metric.  

Fig. \ref{fig:XXK0VJARWR_convergence_metrics_over_generations_all_pop}.(a) and \ref{fig:XXK0VJARWR_convergence_metrics_over_generations_all_pop}.(b) depict the convergence of $\ErrorData$. By using the multi-objective function (m.o.f.) \textit{sdc} we observe higher variance with respect to the m.o.f. \textit{d} throughout the generations. This is due to the fact that some FTs are Pareto optimal in other aspects, e.g. FTs with a small size, and often these FTs have higher error. In contrast, Fig. \ref{fig:XXK0VJARWR_convergence_metrics_over_generations_all_pop}.(b) shows less variance, which tells that FTs within a generation have a similar error based on the failure data set ($\ErrorData$).

Fig. \ref{fig:XXK0VJARWR_convergence_metrics_over_generations_all_pop}.(c) and \ref{fig:XXK0VJARWR_convergence_metrics_over_generations_all_pop}.(d) depict the convergence of $\ErrorMCSs$. We observe a similar pattern, where our approach keeps more ``variety'' of FTs structures between generations where each of those is to be Pareto efficient in at least one metric. In Fig \ref{fig:XXK0VJARWR_convergence_metrics_over_generations_all_pop}.(d) we observe that variance decreases with the generations, which tells that FTs tend to be more similar in their MCS matrices in latter generations.

Fig. \ref{fig:XXK0VJARWR_convergence_metrics_over_generations_all_pop}.(e) and \ref{fig:XXK0VJARWR_convergence_metrics_over_generations_all_pop}.(f) depict the convergence of $\FTSize$. Here we can see the most significant differences between both approaches. We observe in Fig. \ref{fig:XXK0VJARWR_convergence_metrics_over_generations_all_pop}.(e) that the FTs tend to be small. It is important to notice that right before finding the global optimum, the FTs within a generation tend to increase in size. Once the global optimum is found (i.e., $\ErrorData = \ErrorMCSs = 0$), the remaining part of the process, naturally (since the error of the best FT candidate(s) cannot be minimised anymore) is related to minimise the FT size, which yields a \textit{compressed} version of the found global optimum. On the other hand, Fig. \ref{fig:XXK0VJARWR_convergence_metrics_over_generations_all_pop}.(f) shows that not controlling the size of the FT results in \textit{structural explosion}, yielding massive structures that are not beneficial for the inference process.

\section{Comparing the performance of m.o.f.'s for the case studies CSD, PT, and SMS'}\label{app:comp_performance_cases_CSD_PT_SMS}

\begin{figure}[!h]
\centering
\begin{subfigure}{.23\textwidth}
  \centering
  \includegraphics[width=1\linewidth]{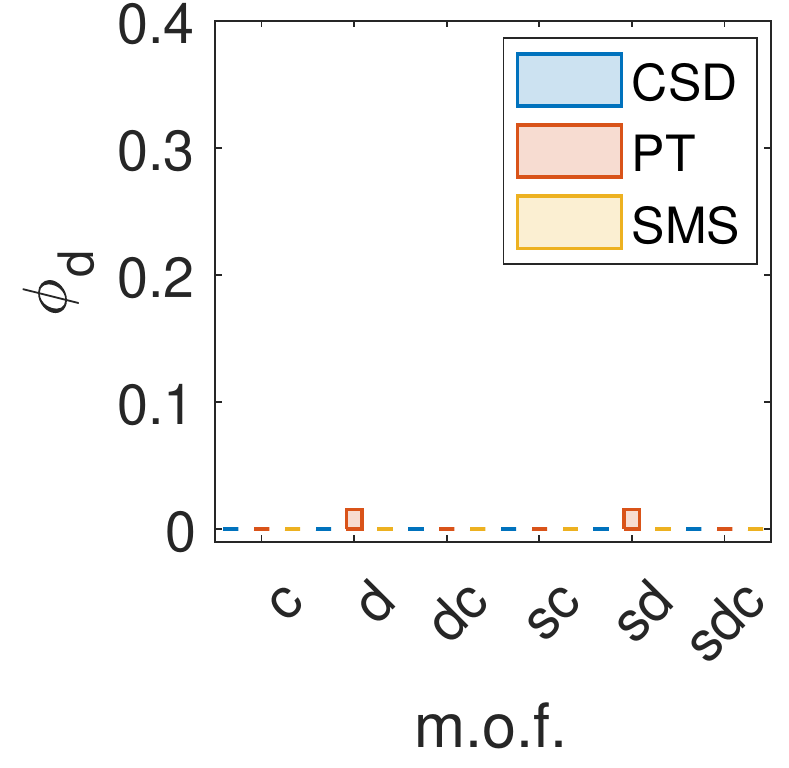}
  \caption{}
\end{subfigure}
\begin{subfigure}{.23\textwidth}
  \centering
  \includegraphics[width=1\linewidth]{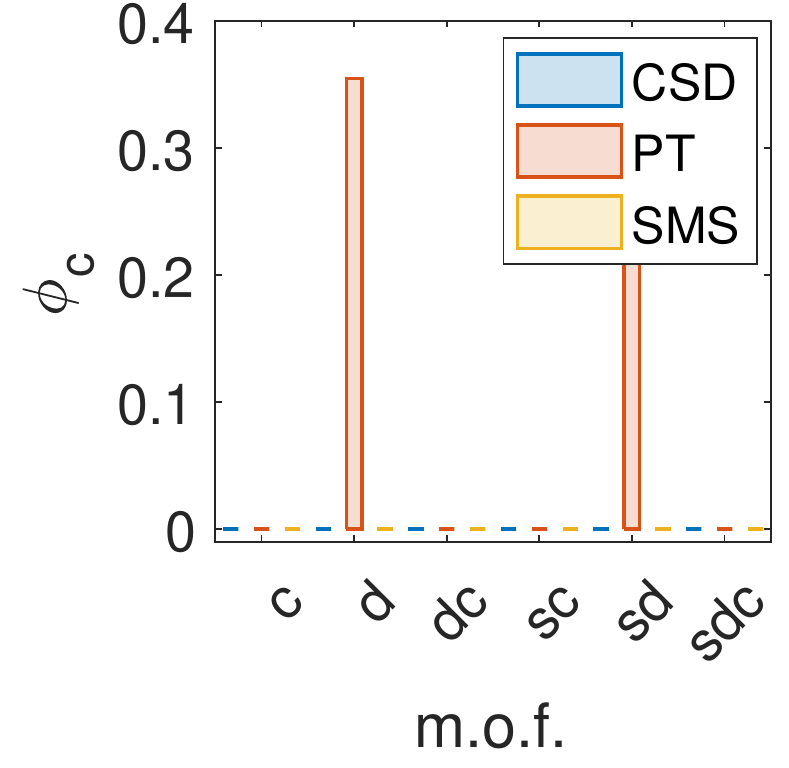}
  \caption{}
\end{subfigure}

\begin{subfigure}{.23\textwidth}
  \centering
  \includegraphics[width=1\linewidth]{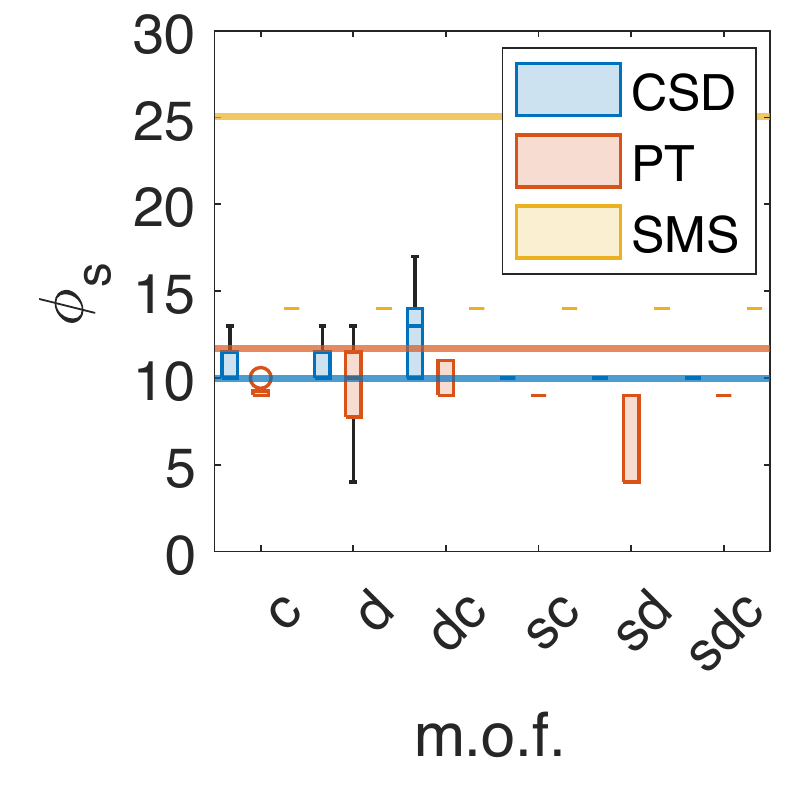}
  \caption{}
\end{subfigure}%
\begin{subfigure}{.23\textwidth}
  \centering
  \includegraphics[width=1\linewidth]{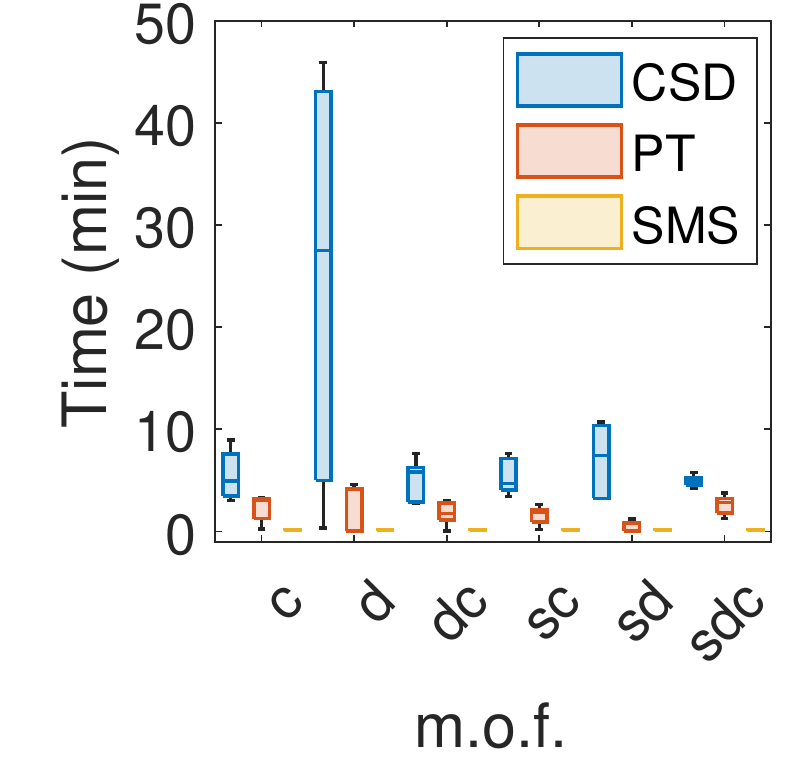}
  \caption{}
\end{subfigure}%
\caption{Comparing the performance of all m.o.f.'s using $ps=400$, $ng=100$, $uc=20$ and the case studies CSD, PT, and SMS. In (a) $\ErrorData$, (b) $\ErrorMCSs$, (c) $\FTSize$, (d) convergence time.}
\label{fig:KCaYZmwRj4_comparing_objective_functions2}
\end{figure}

Fig. \ref{fig:KCaYZmwRj4_comparing_objective_functions2} compares the performance of multi-objective functions (m.o.f.'s) for the case studies CSD, PT, and SMS in Table \ref{tb:case_studies}, using $ps=400$, $ng=100$, $uc=20$.

\section{Data-driven methods to infer FTs from data}\label{app:ddFTs_table}

Table \ref{tb:state_of_the_art_literature} (divided in two parts) contain references associated to data-driven algorithms to infer FTs, the name of the algorithm (if any), whether it is publicly available (if `yes' the table provides a hyperlink redirecting to the respective online repository), the key aspects of the methodology, the input data, the benefits, and drawbacks.

\begin{table*}[ht]\setlength\tabcolsep{4pt} 
\centering
\small
\caption{(Part I) \textit{Data-driven} approaches for the automatic inference of FT models. In the columns the references, algorithm name, whether the algorithm is publicly available or not, the key aspects of the methodology, the input data \{Time Series (TS), Binary Data (BD), Text, System Domain and Relations (SDR)\}, the benefits and drawbacks.}
 \begin{tabular}{p{0.10\textwidth}p{0.06\textwidth}p{0.05\textwidth}p{0.17\textwidth}p{0.05\textwidth}p{0.22\textwidth}p{0.22\textwidth}}
    \toprule
    \textit{Reference(s)} & \textit{Name}  & \textit{Aval?} & \textit{Methodology} & \textit{Input} &  \textit{Benefits} & \textit{Drawbacks} \\ \midrule
    
    Madden \& Nolan (1993) \cite{madden1970generation}, (1998) \cite{madden1970hierarchically}, (1999) \cite{madden1999monitoring} & IFT & No & Based on the ID3 algorithm \cite{quinlan1986induction} to induce DTs. & TS & Purely data-driven approach. Provides insights to carry out rules-based diagnostics. &  It is unclear whether the IFT algorithm guarantees the encoding of cause-effect relationships, which is a requirement for FT models. \\
    
    \hline
    
    Berikov (2004) \cite{berikov2004fault} & - & No & They build FT models based on DTs. & TS & They build the FT model in layers from the DT, this is an interesting way to exploit DTs capabilities. & It is difficult to determine automatically the non-terminal nodes of the FT without expert advice.\\
    
    \hline
    
    Mukherjee \& Chakraborty (2007) \cite{mukherjee2007automated} & - & No & Based on text mining and natural language processing. & Text & Makes use of data such as maintenance reports. & It requires a manually built (partial) FT, which is then refined through their method. \\
    
    \hline
    
    Roth et al. (2015) \cite{roth2015integrated} & - & No & Based on various matrix analysis methods commonly used to model dependencies. & SDR & The approach helps to identify critical failures or elements from a structural point of view. Also enables handling alternative input data. & It requires functional analysis and expert domain support. \\
    
    \hline
    
    Nauta et al. (2018) \cite{nauta2018lift} & LIFT & \href{https://github.com/M-Nauta/LIFT}{Yes} & Based on the Mantel-Haenszel statistical test. & BD & It works well for small problems and with low noise levels in the data. &  It requires as input information about the intermediate events. Moreover, the algorithm does an exhaustive search, and thus has exponential time complexity.\\ 
    
    \hline
    
    Waghen \& Ouali (2019) \cite{waghen2019interpretable} & ILTA & No & Knowledge Discovery in Data set (KDD) + Interpretable Logic Trees (ILT) & TS & It does not require human expertise in the construction stage. It generates intuitive logic trees structures that encode hidden system’s causal relations. & The ILTA algorithm is insufficient when several subsystems and characteristic variables may have interdependent relationships with each other.\\
    
    \bottomrule
 \end{tabular}\label{tb:state_of_the_art_literature}
\end{table*}

\begin{table*}[ht]\setlength\tabcolsep{4pt} 
\centering
\small
\ContinuedFloat
\caption{(Part II) \textit{Data-driven} approaches for the automatic inference of FT models. In the columns the references, algorithm name, whether the algorithm is publicly available or not, the key aspects of the methodology, the input data \{Time Series (TS), Binary Data (BD), Text, System Domain and Relations (SDR)\}, the benefits and drawbacks.}
 \begin{tabular}{p{0.08\textwidth}p{0.06\textwidth}p{0.05\textwidth}p{0.19\textwidth}p{0.05\textwidth}p{0.23\textwidth}p{0.23\textwidth}}
    \toprule
    \textit{Reference(s)} & \textit{Name}  & \textit{Aval?} & \textit{Methodology} & \textit{Input} &  \textit{Benefits} & \textit{Drawbacks} \\ \midrule
    
    Linard et al. (2019) \cite{linard2019fault} & FT-EA & \href{https://gitlab.science.ru.nl/alinard/learning-ft}{Yes} & Based  on  evolutionary algorithms and a one-dimensional cost function based on what the authors call accuracy computed from the labelled binary fault data set. & BD & It manages to achieve FT models with a small error based on the fault data set. Additionally, the algorithm handles noisy data. & The cost function does not take into account the size of the FT model, which in large problems can result in massive structures that are difficult to interpret and which increases the computational time to converge to a solution. It can only handle AND and OR gates.\\ 
    
    \hline
    
    Linard et al.(2020) \cite{linard2019induction} & FT-BN & \href{https://gitlab.science.ru.nl/alinard/learning-ft-bn}{Yes} & First learn the parameters of a Bayesian Network (BN) and then translates the BN into a FT. & BD & Their results show that the method is particularly robust handling noisy data. & It requires prior assumptions based on \textit{white-} and \textit{black-listing}, which define arcs that are respectively missing or present in the BN. \\
    
    \hline
    
    Lazarova-Molnaretal (2020) \cite{Feng2020DATADRIVENFT} & DDFTA & No & It makes use of binarization techniques, MCSs, and Boolean algebra & TS & It is efficient in terms of computational time. Moreover, the algorithm infers VoT logic gates. & It is unclear how the algorithm performs under noisy data. \\
    
    \hline
    
    Waghen \& Ouali (2021) \cite{waghen2021multi} & MILTA & No & Uses an iterative burn-and-build algorithm to identify causal relations between the fault event and its intermediate causes, level after level, until the root-causes are uncovered. & TS & The algorithm does not require the participation of any human experts to build the model. In addition, this method focuses on finding patterns in the data that may indicate the logical relationships of the FT model. & The authors mention that when the system undergoes degradation over time, the MILTA model needs to be improved to capture the evolution of the failure event over time. \\
    
    \hline

    \textit{This paper} & FT-MOEA & \href{https://gitlab.utwente.nl/jimenezroala/ft-moea}{Yes}  & Multi-objective evolutionary algorithms based on the Non-dominated Sorting Genetic Algorithms and Crowding-Distance & BD & The multi-target function is highly customizable. By minimizing the size of the FT model, it provides small and easy-to-interpret structures, which can be considered a compression technique. Moreover, they often converge to the same FT structure. & The computation time increases exponentially with the complexity of the problem. It can only handle labeled binary data, and infers FT models using only AND and OR gates. Depending on the cost function, it can fall into local optimum. \\ \bottomrule
 \end{tabular}
\end{table*}

\end{document}